\documentclass[journal]{IEEEtran}
\usepackage{cite}

\ifCLASSINFOpdf
   \usepackage[pdftex]{graphicx}
\else
\fi
\usepackage{amsmath}
\usepackage{setspace}

\usepackage{soul} 
\usepackage{booktabs} 

\usepackage[colorlinks=true,linkcolor=blue,citecolor=blue,urlcolor=black ]{hyperref}%

\usepackage{graphicx}
\usepackage{algorithm,algpseudocode}
\algnewcommand{\Inputs}[1]{%
\State \textbf{Inputs:}
\Statex \hspace*{\algorithmicindent}\parbox[t]{\linewidth}{\raggedright #1}
}
\algnewcommand{\Initialize}[1]{%
\State \textbf{Initialize:}
\Statex \hspace*{\algorithmicindent}\parbox[t]{\linewidth}{\raggedright #1}
}

\usepackage[square,numbers]{natbib}
\usepackage{mathtools}
\usepackage[labelformat=simple]{subfig}

\usepackage{multirow}
\usepackage{amssymb}

\usepackage{cite}

\begin{document}
%
\title{A Cost-Effective Person-Following System for Assistive Unmanned Vehicles with Deep Learning at the Edge}
%
%
%

\author{Anna~Boschi,
        Francesco~Salvetti,
        Vittorio~Mazzia,
        and~Marcello~Chiaberge
\thanks{The authors are with Politecnico di Torino -- Department of Electronics and Telecommunications Engineering, Italy. Email: \{name.surname\}@polito.it.}}

%
%

\markboth{}%
{Anna Boschi \MakeLowercase{\textit{et al.}}: A Cost-Effective Person-Following System for Assistive Unmanned Vehicles with Deep Learning at the Edge}
%



\maketitle

\begin{abstract}
The vital statistics of the last century highlight a sharp increment of the average age of the world population with a consequent growth of the number of older people. Service robotics applications have the potentiality to provide systems and tools to support the autonomous and self-sufficient older adults in their houses in everyday life, thereby avoiding the task of monitoring them with third parties. In this context, we propose a cost-effective modular solution to detect and follow a person in an indoor, domestic environment. We exploited the latest advancements in deep learning optimization techniques, and we compared different neural network accelerators to provide a robust and flexible person-following system at the edge. Our proposed cost-effective and power-efficient solution is fully-integrable with pre-existing navigation stacks and creates the foundations for the development of fully-autonomous and self-contained service robotics~applications.
\end{abstract}


%
\IEEEpeerreviewmaketitle

\section{Introduction}

Person-following is a well-known problem in robotic autonomous navigation that consists of the ability to detect and follow a target person with a mobile platform.  This task can be achieved with a~variety of sensing and moving systems and has fundamental roles in a variety of applications in domestic, industrial, underwater and aerial scenarios~\cite{islam2019person}. Due to the sharp increment of life expectancy in the last century, the~world population has seen a progressive increase in the number of older people~\cite{worldpopulationageing2019}. This trend offers an excellent opportunity for developing new service robotics applications to provide continuous assistance to autonomous elders in everyday life. These robotic platforms should be able to identify the target person and follow him to offer their support. Person-following assumes, thus, a vital role, as~a technology necessary to enable a variety of different applications. In~this context, it is essential to develop a system that focuses on robustness to different domestic scenarios and efficiency to be implemented on low-power devices, without~the need for external computing devices. Moreover, the~ability to run a person-following algorithm entirely onboard makes the system less prone to security and privacy issues, avoiding unnecessary transmission of sensitive information, such as domestic camera~streams.

Generally speaking, a~person-following system is composed of a sensing device, a~detection algorithm able to provide an estimate of the target position and a following algorithm to control the robot's movements. Indoor robotic platforms use a variety of perception devices, divided into exteroceptive, such as cameras, LiDARs and ultrasonic sensors, and~proprioceptive, such as inertial measurement units (IMUs), gyroscopes, accelerometers and encoders. Different solutions to the detection problem can be found in the literature, depending on the used sensors, on~the application scenario and the type of approach~\cite{islam2019person}. Recent developments in deep learning techniques~\cite{lecun2015deep} for computer vision gave a significant boost to the ability to efficiently extract meaning from visual information and inspired several solutions for the person-following~problem.

Inspired by these approaches, we employed the popular deep learning object detection algorithm YOLOv3-tiny~\cite{redmon2018yolov3}, suitably re-trained for the specific task, to~detect the target person from an RGB-D frame and compute his location with respect to the robot reference frame. The~extracted information was then fed to an efficient control algorithm that generated the suitable linear and angular commands for the robot actuators to achieve person-following. Moreover, we tested the proposed approach on several embedded platforms designed explicitly for the {{edge AI}}, that consists of deploying artificial intelligence algorithms on low-power devices. We compared the obtained results with a particular focus on the trade-off between performance and power consumption. The~overall  solution proposed represents a cost-effective, low-power pipeline for the person-following problem that can be easily employed at the edge as a primary component in complex service robotics~tasks.

\section{Related~Works}
Related literature is organized as follows. Firstly, several methods for  person-following are analyzed, with~a focus on the sensing devices used and on the strategies used to detect the target. Then,~deep~learning techniques for object detection are briefly discussed, with~attention paid to recent developments in edge AI~implementations.

\subsection{Person~Following}
The task of recognizing and localizing a person to be followed by a robotic platform has been widely discussed in the literature since the nineties. Islam~et~al.~\cite{islam2019person} reviewed and categorized a large number of works focused on achieving person-following in a variety of conditions, such as ground, underwater and aerial scenarios. For~what concerns terrestrial applications, important classifications of the different methods are based on the kinds of devices used to sense the environment and on the strategy used to detect the target~person.

Most ground applications use a simple unicycle model that controls the robot 2D motion in polar coordinates, with~a linear velocity on the xy plane and angular velocity about the z-axis~\cite{pucci2013nonlinear}. The~chosen detection system should, therefore, be able to find the target position and distance from the robot. Several systems use laser range finders (LRF) measures that directly provide a set of distances, which are clustered and interpreted to extract relevant features. People are localized usually by means of leg~\cite{chung2011detection,morales2012people,cosgun2013autonomous,leigh2015person,adiwahono2017human,cen2019real} or torso identification~\cite{jung2012control,cai2014human,koide2016identification}. However, these methods mainly rely on static features extracted from 2D point clouds that frequently lead to a poor detection quality. Visual sensors are much more informative since they allow one to sense the entire body of the target, but~simple RGB cameras are not enough since a distance measure is also needed. The~two main categories of visual sensors able to catch depth information are stereo and RGB-D cameras. Several works~\cite{brookshire2010person,satake2012sift,satake2013visual,chen2017integrating,chen2017person,wang2018person} use the first approach to approximate the distance information by triangulation methods applied on two or more RGB views of the same scene. However, the~most used visual sensors for person detection are RGB-D cameras~\cite{doisy2012adaptive,basso2013fast,munaro2013software,do2015embedded,ren2016real,mi2016system,gupta2016novel,masuzawa2017development,chi2017gait,jiang2018classification,chen2018folo,yang2019control} that are able to get both RGB images and depth maps by exploiting infrared light. Several methods employ sensor fusion techniques to merge information from different kinds of sensing systems. For~example, Alvarez~et~al.~\cite{alvarez2012feature} used both images to detect the human torso and lasers to track the legs, Susperregi~et~al.~\cite{susperregi2013rgb} used an RGB-D camera, lasers and a thermal sensor; and~Wang~et~al.~\cite{wang2017real} used a~monocular camera with an ultrasonic sensor. Hu~et~al.~\cite{hu2013design} adopt eda human walking model using a combination of RGB-D data, LRF leg tracking and robot odometry and a sonar sensor for obstacle avoidance during navigation. Koide~et~al.~\cite{koide2016identification} used LRF data to detect people in the scene, and then cameras to identify them and extract relevant features. Cosgun~et~al.~\cite{cosgun2013autonomous}, on~the other hand, manually selected the target from an RGB-D view of the environment, and then tracked it with LRF leg identification. Merging data from multiple sensors allows one to increase detection accuracy, but~with high increases in the system's complexity and costs. Furthermore, the~presence of multiple sources of data requires hardware with high computational power to enable real-time processing. Since our focus was on developing an embedded, cost-effective, low-power system, we~selected a low-cost RGB-D camera as the only sensing~device.

Focusing on vision-based methods, different strategies can be adopted to detect the person in the environment. Mi~et~al.~\cite{mi2016system}, Ren~et~al.~\cite{ren2016real} and Chi~et~al.~\cite{chi2017gait} all adopted the Microsoft Kinect SDK that directly provides skeleton position. Satake~et~al.~\cite{satake2012sift,satake2013visual} used manually designed templates to extract relevant features and find the target location. Munaro~et~al.~\cite{munaro2013software}, Brookshire~\cite{brookshire2010person} and Basso~et~al.~\cite{basso2013fast}, instead, adopted histograms of the oriented gradients (HOG) method for human detection originally proposed by Dalal~et~al.~\cite{dalal2005histograms}. More recently, machine learning techniques have been used to solve the person-following task. Chen~et~al.~\cite{chen2017person} used an online AdaBoost classifier initialized on a~manually-selected bounding box of the target person. Chen~\cite{chen2018folo}, instead, used an upper-body detector based on an SVM to get the human position and extract relevant features used during the tracking phase. More recently, deep learning models have been employed to further boost detection accuracy. Chen~et~al.~\cite{chen2017integrating} proposed a CNN-based classifier trained on a manually-selected target with an online learning procedure. Masuzawa~et~al.~\cite{masuzawa2017development} adopted the YOLO method~\cite{redmon2016you} to identify the person due to its high results in both precision and recall rates. Wang~et~al.~\cite{wang2018person} also employed YOLO as a person detector, but~only to predict the initial position of the target, since they are not able to run the algorithm in real-time due to hardware limitations. Jiang~et~al.~\cite{jiang2018classification} jointly used a DCNN-based detector and a PN classifier based on random forests to enhance person localization and tracking. Finally, Yang~et~al.~\cite{yang2019control} used a DNN to identify a bounding box image to be scored against the pre-registered user~image.

Exactly as in~\cite{wang2018person,masuzawa2017development}, we purpose a person-following approach based on the YOLO network, but~our methodology is different from theirs. We use a newer and smaller version of YOLO (YOLOv3-tiny), and~the re-training and the optimization of the network involve:
\begin{itemize}
\item[-] Eliminating the tracking part and relating an additional filter thanks to the continuous detection of the target, so reducing the computational complexity of the solution;
\item[-] Running the detection at the edge, so it can be easily realized on the neural board accelerator, without~adding an expensive onboard computer (low-cost).
\end{itemize}

\subsection{Deep Learning for Real-Time Object~Detection}
Object detection is a field of computer vision that deals with localizing and labeling objects inside an image. Before~the recent huge developments in deep learning techniques, object detection was classically performed with machine learning methods such as the cascade classifier based on Haar-like features~\cite{viola2001rapid} or coupled with feature extraction algorithms like the histograms of oriented gradients (HOG) \cite{dalal2005histograms,felzenszwalb2008discriminatively}. With~the recent developments in deep learning, considerable improvements in both the accuracy and efficiency of object detection algorithms have been achieved. Current techniques are split into region proposal methods and single-shot detectors.
The former firstly identifies areas inside the image that most likely contain objects of interest, abd then feeds them to a second stage that predicts label and bounding box dimensions. In~this category, we find algorithms such as R-CNN~\cite{girshick2014rich}, fast~R-CNN~\cite{girshick2015fast} and faster R-CNN~\cite{ren2015faster}. Single-shot detectors, on~the other hand, treat~the detection task as a regression problem and directly perform both localization and labeling with a single stage. This~method makes them generally faster than region proposal techniques, but~with slightly less accuracy. The~most known single shot detectors are SSD~\cite{Liu-2016} and YOLO~\cite{redmon2016you}, with~its evolutions YOLOv2~\cite{redmon2017yolo9000}, YOLOv3~\cite{redmon2018yolov3} and YOLOv4~\cite{bochkovskiy2020yolov4}. Lightweight versions of these methods, such as YOLOv3-tiny, have been specifically developed to be implemented in low-power real-time systems and therefore are most suitable for service robotics~applications.

Recently, several advancements have been made in edge AI, where deep neural networks are deployed on low-power real-time embedded systems~\cite{mittal2019survey}. This field of research has principally flourished thanks to the release of hardware platforms specifically designed to accelerate deep neural network inferences. NVIDIA released boards with onboard GPUs, such as Jetson TX2, AGX Xavier and Nano. Intel produced two generations of USB hardware accelerators called Neural Computing Stick (NCS), and recently Google released its own Coral board and USB accelerator, able to boost inference performance using the Tensor Processing Unit (TPU) chips. In~the literature can be found several works which apply optimization techniques to object detection algorithms to deploy them on embedded devices with hardware acceleration~\cite{xu2017classify,kang2018joint,cao2018detecting,yang2018hybrid,mazzia2020real}.

In our work, we fine-tuned a pre-trained YOLOv3-tiny network for the person detection task, and~we propose a cost-effective person-following system that can generate suitable velocity commands for the robotic actuators, based on RGB-D images. The~proposed methodology was extensively tested with several edge AI devices in order to compare performance and power consumption for the different possible configurations. Finally, a~potential implementation of the proposed system was integrated and tested with a standard robotic~platform.

The rest of the paper is organized as follows.~Section~\ref{sec:materials} presents the dataset used in the re-training and the hardware setup. Section~\ref{sec:methodology} discusses the proposed methodology with an extensive description of the detection mechanism and the control algorithm. Finally, Section~\ref{sec:experimental} presents the experimental discussion, the~performance comparison on the considered hardware platforms and the final complete~implementation.

\section{Materials and~Data}
\label{sec:materials}
The network adopted was pre-trained with the COCO dataset, which contains 80 classes of objects with their respectively bounding box and marks. Subsequently, a~technique named {{\textit{transfer learning}}}~\cite{long2016deep} was adopted to realize the re-training and the fine-tuning of the network using a smaller dataset composed by the {{\textit{person}}} class only. In~this way, a~custom version of the network YOLOv3-tiny was produced, optimizing it for accurate and efficient detection. That network was tested and compared with the original model, producing some metric evaluation results. The~deep learning network was evaluated on different edge AI devices by assessing the performance of each of them in terms of inference speed and power consumption. Finally, a~specific hardware solution was selected to assemble a robotic platform and test it in a real~environment.

\subsection{Data~Description}
The images of people used to create the person dataset were extracted from the OIDv4~\cite{kuznetsova2018open} dataset, which are divided into training, validation and testing. During~the re-training phase, we used 6001~images, divided into the training set, 5401, and~the test set, 600. The~dimension of the images was imposed to be equal to 416 {$\times$} 416 during training, in~order to be more coherent with the pre-trained input dimension of the original network and the native resolution, 480 {$\times$} 640, of~the depth~camera.

\subsection{Hardware~Description}
\label{subsection:S3_2}

The main request to fulfill is the necessity of a real-time response in each step that makes up this robotic application: person detection algorithm, data elaboration, control of the robot and the navigation into the indoor environment. All these operations must be as instantaneous as possible to avoid the loss of the person to follow---extremely probable in case the person moves away from the robot~view.

\begin{figure}[H]
\centering
\includegraphics[scale=0.35]{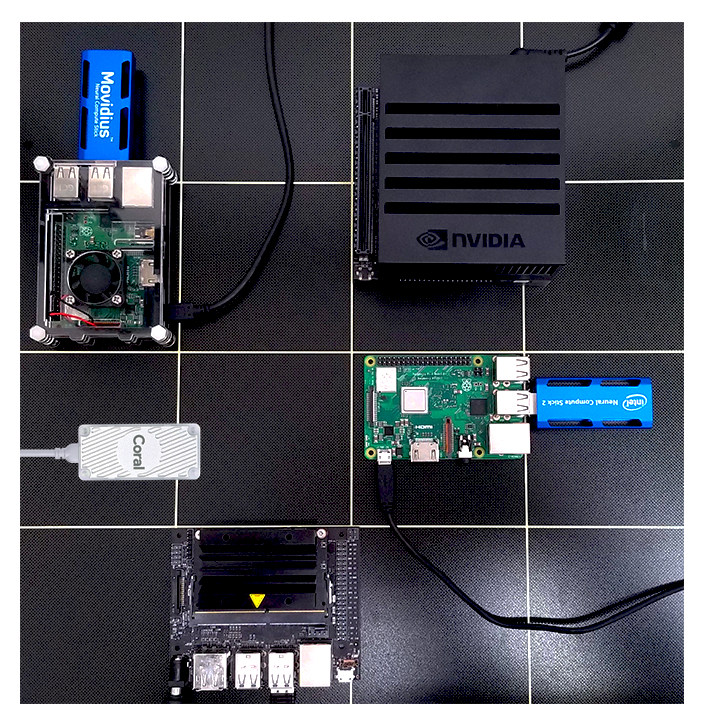}
\caption{The analyzed embedded devices to deploy deep neural networks at the edge. \textbf{Top left}: Raspberry Pi 3 B+ with Intel NCS. \textbf{Top right}: NVIDIA Jetson AGX Xavier. \textbf{Center left}: Coral USB Accelerator. \textbf{Center right}: Raspberry Pi 3 B+ with Movidius NCS2. \textbf{Bottom}: NVIDIA Jetson Nano.}
\label{fig:edge_ai_devices}
\end{figure}

For what concerns the embedded implementation of the neural network, different platforms were evaluated and are shown in Figure~\ref{fig:edge_ai_devices} and summarized in  Table~\ref{tab:edgeAIcomponents}: a Raspberry Pi 3 B+ with Intel NCS, a~Raspberry Pi 3 B+ with Movidius NCS2, a~Coral USB Accelerator, an~NVIDIA Jetson AGX Xavier developer kit and an NVIDIA Jetson~Nano.

The Neural Computing Sticks (NCS) are USB dedicated hardware accelerators specifically used to perform deep neural network inferences. Both the first and the second generations of the NCS have been tested: the first has a Myriad 2 Vision Processing Unit (VPU), while the second has Myriad X VPU and reaches eight times  the performance of the previous version. These two components request a USB 3.0 or 2.0 interface so that they can be easily used with cheap single-board computers such as a Raspberry~board.

\begin{table*}[t]
\centering
\caption{{The main specifications} of the embedded HW for edge AI taken into account for the experimentation.~The~price indicated for each device is referred to the commercial value at the time of the~publication.}

\begin{tabular}{llllll}
\toprule
& \textbf{Intel NCS} & \textbf{\begin{tabular}[c]{@{}l@{}}Intel Movidius \\ NCS2\end{tabular}} & \textbf{\begin{tabular}[c]{@{}l@{}}Coral USB   \\ Accelerator\end{tabular}} & \textbf{\begin{tabular}[c]{@{}l@{}}NVIDIA Jetson \\ AGX  Xavier \\ Developer Kit\end{tabular}} & \textbf{\begin{tabular}[c]{@{}l@{}}NVIDIA \\ Jetson Nano\end{tabular}} \\ \midrule
\begin{tabular}[c]{@{}l@{}}AI \\ performance\end{tabular} & \begin{tabular}[c]{@{}l@{}}100 GFLOPs\\ (FP32)\end{tabular} & \begin{tabular}[c]{@{}l@{}}150 GFLOPs\\ (FP32)\end{tabular} & \begin{tabular}[c]{@{}l@{}}4 TOPs\\ (INT8)\end{tabular} & \begin{tabular}[c]{@{}l@{}}32 TOPs\\ (FP32)\end{tabular} & \begin{tabular}[c]{@{}l@{}}472 GFLOPs\\ (FP32)\end{tabular} \\
\midrule
\begin{tabular}[c]{@{}l@{}}HW \\ accelerator\end{tabular} & \begin{tabular}[c]{@{}l@{}}Myriad 2 \\ VPU\end{tabular} & \begin{tabular}[c]{@{}l@{}}Myriad X \\ VPU\end{tabular} & \begin{tabular}[c]{@{}l@{}}Google \\ Edge TPU \\ coprocessor\end{tabular} & \begin{tabular}[c]{@{}l@{}}512-core NVIDIA \\ Volta GPU with 64 \\ Tensor Cores and \\ 2x NVDLA Engines\end{tabular} & \begin{tabular}[c]{@{}l@{}}128-core NVIDIA \\ Maxwell GPU\end{tabular} \\
\midrule
CPU & N.A. & N.A. & N.A. & \begin{tabular}[c]{@{}l@{}}8-core NVIDIA \\ Carmel Arm \\ v8.2 64-bit CPU \\ 8MB L2 + 4MB L3\end{tabular} & \begin{tabular}[c]{@{}l@{}}Quad-core ARM \\ Cortex-A57 \\ MPCore processor\end{tabular} \\
\midrule
Memory & 4 GB LPDDR3 & 4 GB LPDDR3 & N.A. & \begin{tabular}[c]{@{}l@{}}32 GB 256-bit \\ LPDDR4x \\ 136.5 GB/s\end{tabular} & \begin{tabular}[c]{@{}l@{}}4 GB 64-bit \\ LPDDR4 \\ 25.6 GB/s\end{tabular} \\\midrule
Storage & N.A. & N.A. & N.A. & 32 GB eMMC 5.1 & \begin{tabular}[c]{@{}l@{}}Micro SD card slot \\ or 16 GB eMMC \\ 5.1 flash\end{tabular} \\
\midrule
Power & 1 W & 1.5 W & 1 W & 10/15/30 W & 5/10 W \\\midrule
Size & 73 $\times$ 26 mm & 73 $\times$ 26 mm & 65 $\times$ 30 mm & 100 $\times$ 87 mm & 70 $\times$ 45 mm \\
\midrule
Weight & 18 g & 19 g & 20 g & 280 g & 140 g \\\midrule
Price & \$70 & \$74 & \$60 & \$700 & \$99 \\
\bottomrule
\end{tabular}
\label{tab:edgeAIcomponents}
\end{table*}

The Coral USB Accelerator, released in 2019, is an on-board edge TPU coprocessor able to reach high-performance machine learning inference, with~a limited power cost, for~TensorFlow Lite models. The~board can work at different clock frequencies: maximum or reduced. These frequency types are one the twice of the other,  so using the maximum frequency there is an increase of the inference speed with a consequent increase of the power~consumption.

The NVIDIA Jetson AGX Xavier, released in 2018, is a System-On-Module able to guarantee high performance and power efficiency. The~board contains DRAM, CPU, PMIC, flash memory storage and a dedicated GPU for hardware acceleration, so it has been specifically realized to perform rapidly different neural network operations. The~kit is also supplied with several software libraries as NVIDIA JetPack, DeepStream SDKs, CUDA, cuDNN, and TensorRT. It~is possible to set different power mode configurations also selecting the number of CPU cores utilized: 10 W (2 cores), 15 W (4 cores), 30 W (2,~4, 6 or 8 cores).

The NVIDIA Jetson Nano is a lightweight, powerful computer explicitly designed for AI in order to run multiple neural networks in parallel for image elaboration. The~board mounts a 128-core NVIDIA Maxwell GPU, a~Quad-Core ARM Cortex-A57 MPCore CPU and a 4 GB LPDDR4 memory and reaches the peak performance of 472 GFLOPs. It can work in two power modes: at 5 W or 10 W without the support of Tensor cores during the inference~acceleration.

The complete hardware selected for testing in the test environment is an upgrade of the TurtleBot3 Waffle Pi from  {ROBOTIS}\footnote{\url{https://emanual.robotis.com/docs/en/platform/turtlebot3/specifications/}}, a~standard robotic platform supported by ROS and extremely used by developers. This~robot uses as on-board PC the {Jetson Xavier developer kit}\footnote{\url{https://developer.nvidia.com/embedded/jetson-agx-xavier-developer-kit}}, and~it has been equipped with an additional camera sensor, an~Intel RealSense Depth {camera D435i}\footnote{\url{https://www.intelrealsense.com/depth-camera-d435i/}}.~This low-cost depth camera, ideal for navigation or object recognition applications, is~composed of an RGB module and two infrared cameras separated by a wide IR~projector.

\section{Proposed~Methodology}
\label{sec:methodology}
The goal of this research was to develop an autonomous, real-time, person-following assistive system able to promote the aging-in-place of  independent elderly~people.

The workflow of our solution is  exposed here.~Firstly, the~detection and localization of the person in the environment are realized, using the RGB-D information from the camera and a~neural network specifically designed for fast and accurate real-time object detection, YOLOv3-tiny. Successively,~the~information about the position of the person with respect to the robot reference frame is used to create a tailored control algorithm, using a linear trend for the linear velocities and a~parabolic trend for the angular velocities. In~this way, the~robot can follow the person while remaining at a~certain safe distance from him, thereby avoiding both hitting and losing the target. The~pseudo-code of the overall algorithm is reported in the Algorithm 1.

The proposed algorithm, in~the practical implementation, is integrated with the open-source Robot Operating System (ROS)\footnote{\url{https://www.ros.org/}} to set the control to the actuators of the robot. The~result is an autonomous, cost-effective person-following system with deep learning at the edge, easily integrable in different unmanned~vehicles.
\noindent

\begin{table}[H]
\centering
\begin{tabular}{l}
\toprule
\textbf{Algorithm 1:} Person-following using an RGB-D camera and YOLOv3-tiny~network. \\
\midrule
1: \textbf{Inputs:}\\
\hspace{2em}$D_{h \times w}$: Depth Matrix obtained from the camera \\
\hspace{2em}$x_{bb}$: Horizontal coordinate of the center of the bounding box\\
\hspace{2em}$y_{bb}$: Vertical coordinate of the center of the bounding box \\
\hspace{2em}$x_c$: Horizontal coordinate of the center of the camera frame \\
\hspace{2em}$N_{detection}$: Number of the detections provided by the network\\
2: \textbf{Initialize:}\\
\hspace{2em}$F_{moreperson} \gets 0$ \\
\hspace{2em}$V_{linear_{(0)}} \gets 0$\\
\hspace{2em}$V_{angular_{(0)}} \gets 0$\\
\hspace{2em}$T_{time} \gets 0$\\
\hspace{2em}$i \gets 1$\\
3: \textbf{while} True \textbf{do}\\
4: \hspace{2em}\textbf{if} ($N_{detection}$ == 0) \textbf{then}\\
5: \hspace{4em}\strut$V_{linear_{(i)}} \gets V_{linear_{(i-1)}}$\\
6: \hspace{4em}\strut$V_{angular_{(i)}} \gets V_{angular_{(i-1)}}$\\
7: \hspace{4em}\textbf{if} $T_{time} > t$ \textbf{then}\\
8: \hspace{6em}\textbf{reset:} $T_{time}$\\
9: \hspace{6em}\strut$V_{linear_{(i)}} \gets 0$\\
10:\hspace{6em}\strut$V_{angular_{(i)}} \gets 0$\\
11:\hspace{4em}\textbf{end if }\\
12:\hspace{2em}\textbf{else if} ($N_{detection}$ == 1) \textbf{then}\\
13:\hspace{4em} \strut$dx \gets (x_c - x_{bb})$\\
14:\hspace{4em} \strut$depth \gets D[y_{bb},x_{bb}]$\\
15:\hspace{4em} \strut$V_{linear_{(i)}} \gets linearvelocity(depth)$\\
16:\hspace{4em}\strut$V_{angular_{(i)}} \gets angularvelocity(dx)$\\
17:\hspace{4em} \textbf{reset:} $T_{time}, N_{detection}$\\
18:\hspace{2em}\textbf{else} \\
19:\hspace{4em}\textbf{if} $T_{time} > 1$\\
20:\hspace{6em}\textbf{reset:} $T_{time}$\\
21:\hspace{6em} \strut$V_{linear_{(i)}} \gets 0$\\
22:\hspace{6em} \strut$V_{angular_{(i)}} \gets 0$\\
23:\hspace{6em}\strut$F_{moreperson} \gets 1$\\
24:\hspace{4em}\textbf{end if}\\
25:\hspace{4em}\textbf{reset:} $N_{detection}$\\
26:\hspace{2em}\textbf{end if} \\
27:\hspace{2em}$VelocityController(V_{linear_{(i)}}$, $V_{angular_{(i)}})$\\
28:\hspace{2em}\textbf{if} ($F_{moreperson}$ == 1) \textbf{then}\\
29:\hspace{4em}\textbf{stop} the algorithm for a prefixed time\\
30:\hspace{4em}\textbf{reset:} $F_{moreperson}$, $T_{time}$\\
31:\hspace{2em}\textbf{end if}\\
32:\hspace{2em}$i ++$\\
33:\hspace{2em}\textbf{acquire} next Inputs\\
34:\textbf{end while}\\
\bottomrule
\end{tabular}
\end{table}

\subsection{Person~Localization}
By using a re-trained version of YOLOv3-tiny for object detection and the RGB-D camera chosen for this application, it is possible to detect and localize a person in space interactively.~In~fact, once~the network is optimized, the~precision and recall values allow one to have a continuous detection of the target without the use of a tracker algorithm or additional filter to support the control implementation. That means  the use of the network is sufficient to realize real-time person-following, while reducing the computational cost and other power consumption. These considerations are supported both from the {average precision $AP_{50}$} obtained from the re-trained network, and the specific use case taking into account: a self-sufficient older person in his or her home environment. As~the target to follow is an elderly person, its moving velocity is reduced so the tracker is superfluous and consequently the control is smooth, and~this is also supported thanks to the reduced speed of the~robot. 

Object detection is an important area of research, interested in the processing of images and videos to detect and recognize objects. You only look once (YOLO) \cite{redmon2016you,redmon2017yolo9000,redmon2018yolov3} is the object detection method commonly used in the real-time processing image applications. This model, based on a feed-forward convolutional neural network, is considered an evolution of the single-shot-multibox detector (SSD) concept with the idea of predicting both the bounding boxes and the class detection probability simply analyzing the image once. Its architecture is based on a single neural network trained end-to-end to increase the accuracy and to reduce the predictions of false positives on the~background.

The operations done by the network can be divided into four~steps:
\begin{enumerate}
\item The input image is processed with a grid cell as a reference frame.
\item Each grid cell generates bounding boxes and predicts their confidence rate. The~confidence rate depends on the accuracy of the network during the detection.
\item Each grid cell has a probability score for each class. The~number of classes depends on the dataset used during the training process of the network.
\item The total number of bounding boxes is minimized by setting a minimum confidence rate and using the non-maximum suppression (NMS) algorithm to obtain the final predictions that can be used to generate the final output: an input image with the bounding boxes over the detected objects with the reference classes and the accuracy percentages.
\end{enumerate}

During the evolution of the YOLO architecture, incremental improvements can be recorded in the different versions developed, starting from YOLOv2, which includes many features to increase the performance, until~reaching YOLOv3 and YOLOv4, the~last two versions of the model, in~which there are notable improvements in the capability of the network to detect objects. In~our proposed methodology, we suggest a re-trained and optimized version of YOLOv3-tiny, which is the lightweight version of YOLOv3 with a reduced number of trainable parameters. In~Table~\ref{tab:tinyyolov3} is  the structure of the modified architecture of YOLOv3-tiny  for  only the class {person}.

\begin{table}[H]
\centering
\caption{YOLOv3-tiny architecture designed to work with  only class {person} and an input size of 416~$\times$~416. The~network after training is further optimized to be deployed onboard the robotic~platform.}
\scalebox{1.1}[1.1]{\begin{tabular}{lllll}
\toprule
\textbf{Layer} & \textbf{Type}                 & \textbf{Size/Stride} & \textbf{Filters} & \textbf{Output}         \\\midrule
0     & \textit{{Convolution}} & 3 $\times$ 3/1     & 16      & 416 $\times$ 416 $\times$ 16 \\
1     & \textit{{MaxPooling}}  & 2 $\times$ 2/2     &         & 208 $\times$ 208 $\times$ 16 \\
2     & \textit{{Convolution}} & 3 $\times$ 3/1     & 32      & 208 $\times$ 208 $\times$ 32 \\
3     & \textit{MaxPooling}  & 2 $\times$ 2/2     &         & 104 $\times$ 104 $\times$ 32 \\
4     & \textit{Convolution} & 3 $\times$ 3/1     & 64      & 104 $\times$ 104 $\times$ 64 \\
5     & \textit{MaxPooling}  & 2 $\times$ 2/2     &         & 52 $\times$ 52 $\times$ 64   \\
6     & \textit{Convolution} & 3 $\times$ 3/1     & 128     & 52 $\times$ 52 $\times$ 128  \\
7     & \textit{MaxPooling}  & 2 $\times$ 2/2     &         & 26 $\times$ 26 $\times$ 128  \\
8     & \textit{Convolution} & 3 $\times$ 3/1     & 256     & 26 $\times$ 26 $\times$ 256  \\
9     & \textit{MaxPooling}  & 2 $\times$ 2/2     &         & 13 $\times$ 13 $\times$ 256  \\
10    & \textit{Convolution} & 3 $\times$ 3/1     & 512     & 13 $\times$ 13 $\times$ 512  \\

11    & \textit{MaxPooling}  & 2 $\times$ 2/1     &         & 13 $\times$ 13 $\times$ 512  \\
12    & \textit{Convolution} & 3 $\times$ 3/1     & 1024    & 13 $\times$ 13 $\times$ 1024 \\
13    & \textit{Convolution} & 1 $\times$ 1/1     & 256     & 13 $\times$ 13 $\times$ 256  \\
14    & \textit{Convolution} & 3 $\times$ 3/1     & 512     & 13 $\times$ 13 $\times$ 512  \\
15    & \textit{Convolution} & 1 $\times$ 1/1     & 255     & 13 $\times$ 13 $\times$ 18  \\
16    & \textit{YOLO}        &             &         &                \\
17    & \textit{Route 13}    &             &         &                \\
18    & \textit{Convolution} & 1 $\times$ 1/1     & 128     & 13 $\times$ 13 $\times$ 128  \\
19    & \textit{Up-sampling} & 2 $\times$ 2/1     &         & 26 $\times$ 26 $\times$ 128  \\
20    & \textit{Route 19 8}  &             &         &                \\
21    & \textit{Convolution} & 3 $\times$ 3/1     & 256     & 26 $\times$ 26 $\times$ 256  \\
22    & \textit{Convolution} & 1 $\times$ 1/1     & 255     & 26 $\times$ 26 $\times$ 18  \\
23    & \textit{YOLO}        &             &         &               \\
\bottomrule
\end{tabular}}
\label{tab:tinyyolov3}
\end{table}

\subsubsection{Person Detection and Localization~Implementation}
The RGB camera frames are the input data that feed the re-trained YOLOv3-tiny network for the class {person}. As~already introduced, the~full input image is treated in functional regions represented as a grid of cells. In~each region, bounding boxes are weighted by the predicted probabilities, and~the predictions are the result of single network~evaluation.

\begin{figure}[t]
\centering
\includegraphics[width=.482\textwidth]{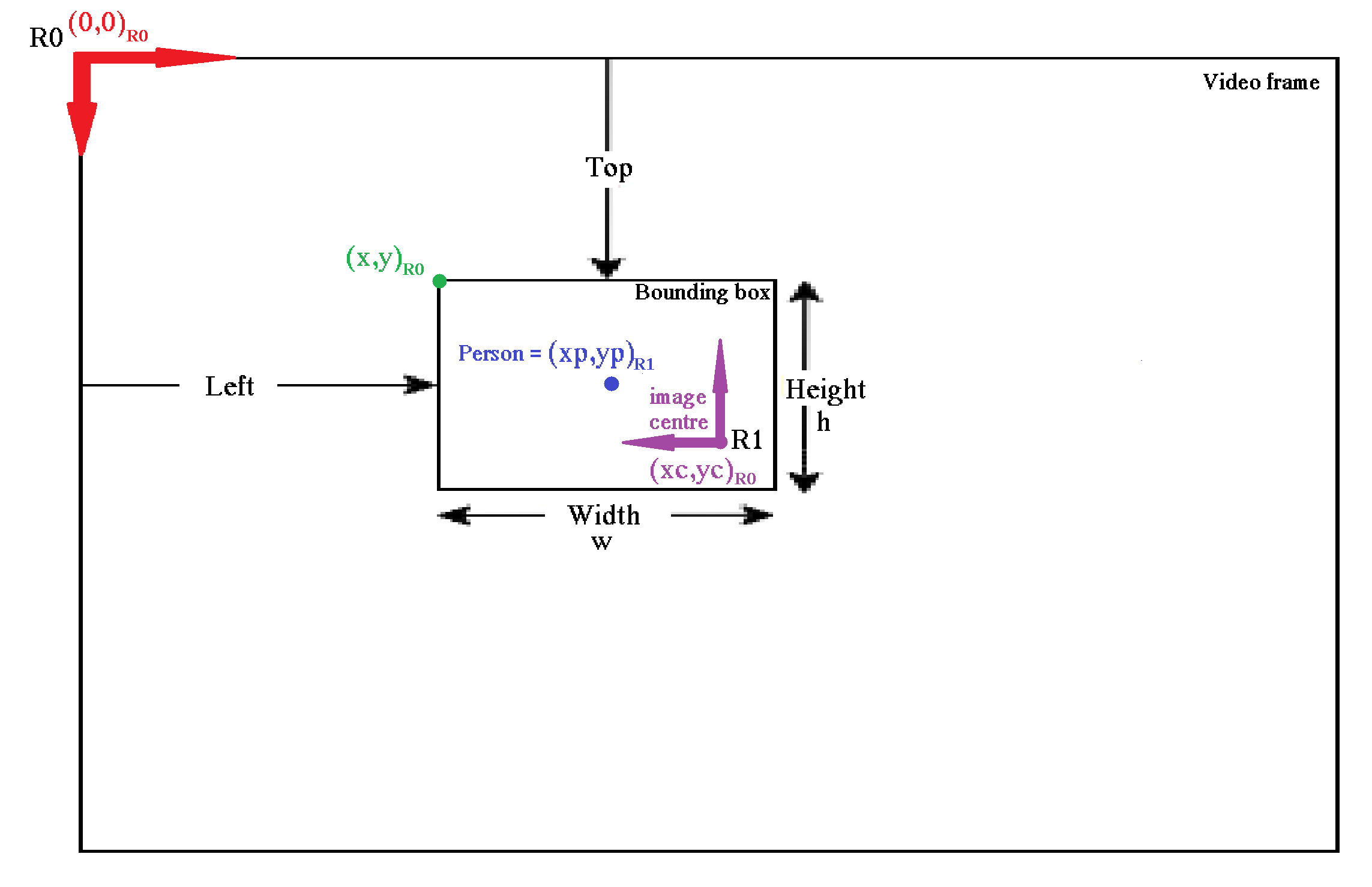}
\caption{Video frame structure. It represents the reference frame (RF) transformation adopted to compute the pixel coordinates of the center of the bounding box detecting the person with respect to a~new RF (R1), the~one used during the calculation of the angular velocity control signals for the~robot.}
\label{fig:bb_referenceframe}
\end{figure}

In order to localize the position of the person in the video frame, it is sufficient to use the bounding box information provided by the network. As~it is depicted in Figure~\ref{fig:bb_referenceframe}, the~bounding box of the detected person directly provides the coordinates of the angle, $x$, $y$, with~respect to the R0 reference~frame.

By using the information of the center, with~respect to R0, it is possible to calculate the coordinates of the midpoint of the person detected with respect to the new reference frame R1 ($x_p$, $y_p$):
\begin{equation}
x_p = x_c - \left(x + \frac{w}{2} \right)
\end{equation}
\begin{equation}
y_p = y_c - \left(y + \frac{h}{2} \right)
\end{equation}
where $x_c$ and $y_c$ are respectively 319 and 239 pixels due to the image resolution, 640 $\times$ 480, of~the camera taken as reference in this~study.

The $x_p$ and $y_p$ coordinates are necessary to locate the person in the 2D space and $x_p$; in~particular, is fundamental to develop the angular control algorithm,  to adjust the rotation of the~robot.

Once known, with the pixel coordinates of the center of the bounding box detecting the person, it is possible to obtain the distance between the robot and the person by merely extracting this information from the corresponding pixel of the depth camera matrix. The~dimension of that matrix is equal to the resolution of the acquired RGB frame, and~for each pixel position, there is a value in millimeters representing the distance of the camera from what it sees. Since the limits of the depth computation, of~the camera taken as reference in this study, are 0.105 m and 8 m, the~values of the depth matrix range from 0 to~8000.

The depth value extracted represents the missing coordinate to localize a person in the 3D environment. The~$z$ coordinate is strictly necessary to realize the linear velocity control algorithm, which is able~to regulate the forward or backward movement of the~robot.

\subsubsection{Detection Situation~Rules}
Three possible detection situations are taken into account by controlling the output value, $N_{detection}$, of~the~network:
\begin{enumerate}
\item \textit{{Nothing detected:}} If nothing is detected, the~robot stops. Differently, suppose the robot loses the person it is following. In~that case, it continues to move in the direction of the last detection with the previous velocity commands for a pre-imposed time $t$. After~that, if~nothing is detected again, the~robot stops.
\item \textit{{One person detected:}} The robot follows the movement of the person while remaining at a certain safe distance from him.
\item \textit{{More than one person simultaneously detected:}} The robot stops for a prefixed time and then it restarts the normal detection operation. There could be many other solutions to implement, for~example, a~person tracking algorithm to follow one of the people detected~\cite{wojke2017simple,Chen2018RealTimeMP,mitzel2011real}. In~particular, the~presence of a tracking algorithm or an additional filter will have to be considered if the use case is changed---for~example, in~case this approach will be used in an office~environment.
\end{enumerate} 

However, for~this specific application, it has been decided to block the robot directly because it has been assumed that the person using it should be self-sufficient, living in the house alone. Therefore,~if~the person  receives visits, it would be unpleasant and unnecessary to have a robot following him inside the house.

\subsection{Person-Following Control~Algorithm}
The linear and the angular velocity are regulated using different functions, so, in~order to obtain a~correct control algorithm, it is necessary to combine them~simultaneously.
\subsubsection{Angular Velocity~Control}
\begin{figure}[H]
\centering
\includegraphics[width=0.480\textwidth]{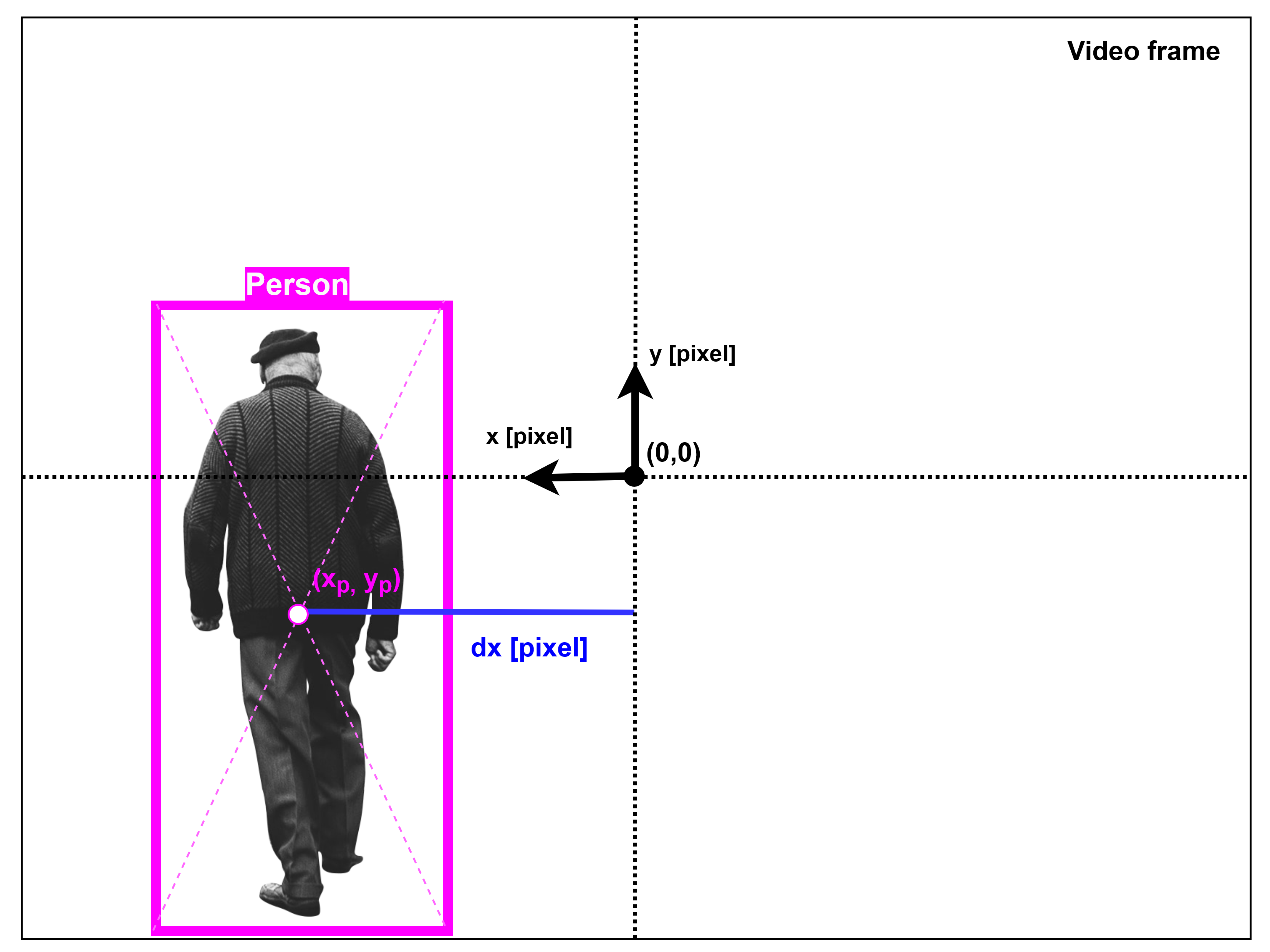}
\caption{The RGB frame of the camera showing how the dx value, adopted in the computation of the angular velocity, is obtained. This value is equal to the horizontal coordinate of the center of the bounding box detecting the person in the~frame.}
\label{fig:video_frame}
\end{figure}

The angular velocity is proportional to the horizontal coordinate, $x_p$, of~the center of the bounding box detecting the person, computed with respect to the reference frame located in the center of the frame of the camera. It has a parabolic trend in order to make the movements of the robot smoother and more natural. By~considering this reference frame, the~$dx$ value is positive if the person is on the left side of the frame or negative if it is on the right side. Figure~\ref{fig:video_frame} shows a graphical representation of how the $dx$ value is~obtained.

The controller function generated has the $dx$ value, measured in pixel, as~input, and gives as output the angular velocity, $v_{angular \theta}$, according to the following formula:
\begin{equation}
v_{angular \theta} =
\begin{cases}
\frac{max_{vel}  \cdot  dx^{2}}{320^{2}} , & \text{ if } dx\ge0 \\
\frac{min_{vel}  \cdot  dx^{2}}{320^{2}} , & \text{ if } dx<0
\end{cases}
\end{equation}

The terms $max_{vel}$ and $min_{vel}$, which are equal and opposite values, are the upper and the lower limits of the angular velocity of the robotic platform. The~number 320 pixels stands for the maximum number of pixels for each side (left and right) of the video frame because the resolution of the image received from our camera is~640 $\times$ 480.

\subsubsection{Linear Velocity~Control}
The linear velocity depends linearly on the distance between the robot and the person detected $d_m$ measured in meters. In~particular, it is in function of the value obtained from the depth matrix of the camera in the center point of the bounding box detecting the person. {This control can be represented by a linear trend divided into three regions.} 
In the first region, the~distance is superior compared to the set upper limit $m_{v_{uplim}}$, so here the robot moves straight on following a linear proportional trend until it reaches its maximum speed saturating to that value.
In~the second region, there is a stop condition; in fact, the~robot is at the safety distance from the person and remains there to avoid losing the person.
The zero value is also assigned when the distance is 0 m, in~order to avoid a particular case in the code. In~fact, the~depth value obtained from our camera has a limit of 0.105 m, so~nothing should be detected at a lower distance.
The final region is between the limit of the camera: 0.105 m and the distance lower-limit $m_{v_{lowlim}}$. In~this condition, the~robot goes back following a linear proportional trend until it reaches its maximum negative speed saturating to that value.
Here is reported the formula responsible for the linear velocity $v_{linear x}$:

\begin{equation}
v_{linear x} =
\begin{cases}
d_m \cdot m1+q1 , \text{ if } d_m > m_{v_{uplim}} \\
0 , \text{ if } m_{v_{lowlim}} < d_m \leq  m_{v_{uplim}} \text{ or } d_m = 0\\
d_m \cdot m2+q2 , \text{ if } d_m \leq m_{v_{lowlim}}
\end{cases}
\end{equation}
\\
The values $m1$, $q1$, $m2$ and $q2$ were found using the equation of the straight line passing through two~points.

\section{Experimental Discussion and~Results}
\label{sec:experimental}
In this section, we firstly discuss some technical details of the re-training procedure of the tiny version of YOLOv3.
Then experimental evaluations are discussed for
both the model and its deployment on the selected embedded devices. Finally, we present our platform implementation that represents one of the possible practical configurations to realize the person-following solution presented with this~work.

\subsection{Person Detector Training and~Optimization}
In order to obtain a lightweight and efficient network for the detection of the target, we modified the original model to classify and localize the class {person} only. Using OIDv4~\cite{kuznetsova2018open}, we collected a set $\mathbb{X}$ of 6001 training samples, reserving 600 of them for testing. Making use of transfer learning~\cite{tan2018survey}, we~started our training from a pre-trained backbone, from~layer 0 to 15 in Table~\ref{tab:tinyyolov3}. That greatly speeds up the training, drastically reducing the number of samples required to achieve a high level of accuracy. We trained for 20 epochs with a linear learning rate decay and an initial value of $\eta=0.0001$. We adopted momentum optimization~\cite{polyak1964some} with $\beta=0.9$ and a batch size of 32. The~training procedure lasted approximately one hour on a workstation with an NVIDIA RTX 2080 Ti and 64 GB of DDR4~SDRAM.

It is possible to observe the effectiveness of the re-training procedure from
Table~\ref{tab:trasnfer_learning}. The~re-trained version of YOLOv3-tiny gains more than 30\% of average precision (AP) at 0.5 of intersection over unit (IOU). Moreover, the~resulting single-class network is 23\% faster, in~terms of inference latency, than~the multi-class counterpart. That is due to the reduced number of features maps in the final detection section of the network, from~layer 15 to 23 of Table~\ref{tab:tinyyolov3}.

\begin{table}[H]
\caption{Average precision at 0.5 IOU for {person} class before and after re-training. It is clear how transfer learning is so effective at improving the performance of the tiny version of the YOLOv3 model, increasing  the metric~score by more than 30\%.}
\centering
\begin{tabular}{lll}
\toprule
\bf{Network}            & \boldmath{$AP_{50}$}       &  \bf{Gain}     \\ \cmidrule{1-3}
YOLOv3-tiny        & 19.21 \% &          \\ \cmidrule{1-3}
YOLOv3-tiny$_{person}$ & 49.30 \% & 30.09 \% \\
\bottomrule
\end{tabular}
\label{tab:trasnfer_learning}
\end{table}
\vspace{-3pt}

Finally, we optimized the resulting re-trained model with two different libraries: TensorRT and TensorFlow Lite\footnote{\url{https://www.tensorflow.org/lite}}. Optimization is a fundamental process and aims at reducing latency, inference cost, accelerator compatibility, memory and storage footprint. That~is mainly achieved with two distinct techniques: model pruning and quantization. The~first one simplifies the topological structure, removing unnecessary parts of the architecture, or~favors a more sparse model introducing zeros to the parameter tensors. On~the other hand, quantization reduces the precision of the numbers used to represent model parameters from float32 to float16 up to int8. That can be accomplished after the training procedure (post-training quantization) or during the training procedure (quantization-aware training), adding fake quantization nodes inside the network and making it robust to quantization noise. Indeed, optimizations can potentially result in changes in model accuracy, and~so any operation must be carefully~evaluated.

In order not to affect the accuracy of our YOLOv3-tiny implementation, we applied basic pruning optimizations with the TensoRT library, removing unnecessary operations and setting to zero irrelevant weights. Indeed, person detection is a critical step in our solution, and~it requires the maintenance of a certain level of performance. Nevertheless, in~order to test also the performance of the custom TPU ASIC of the Coral Accelerator, we produced a full integer model with TensorFlow-Lite optimizer to be compatible with the hardware of the device. When~only applying model pruning we obtained an insignificant accuracy loss; with~8-bit precision the model loses 22\% of its original $AP_{50}$. Indeed, darker scenes, with partially~occluded and small targets, are not precisely detected anymore.~However, latency and inference costs are significantly reduced using this extreme optimization~procedure.

\subsection{Inference with Edge AI~Accelerators}
After the training and optimization procedures, the~re-trained model was deployed on the different edge device configurations presented in Section~\ref{sec:materials}. We tested the performance in terms of absorbed power and frame rate in order to outline different hardware solutions for our proposed cost-effective person-following system. A~single-board computer, Raspberry Pi 3B+, was used for all configurations that require a host~device.

Firstly, we measured the power consumption of the different solutions at an idle condition, and~then we executed the model for approximately five minutes to reach steady-state behavior. We  directly measured the current absorbed from the power source, thereby obtaining the power consumption of the entire~system.

Since the Jetson boards allow the user to select different working power conditions, we tested all of them. The~results are presented in Table~\ref{tab:performance}. The~second version of the Intel Movidius Neural Stick achieves a higher frame rate with less power consumption. However, either Jetson Nano running modes reach higher performance at the expense of higher current absorption. On~the other hand, Jetson AGX Xavier achieves a much higher frame rate on all running modes, but~with other levels of power consumption. Finally, full integer quantization greatly reduces the latency of the model running at more than 30 fps with only 7 W. However, as~previously stated, the~accuracy loss in this last case could compromise the correct functioning of the entire system in certain types of~application.

\begin{table*}[t]
\caption{Comparison between different devices' power consumption levels and performances achieved with the re-trained and modified version of YOLOv3-tiny. The~fps * achieved with the Coral Accelerator are obtained with an int8 weights~precision.}
\centering
\begin{tabular}{llllll}
\toprule
\bf{Device}                  & \textbf{Mode}        & \boldmath{$V_{al}$~{[}$V${]}} & \boldmath{$I_{mean}$~{[}$A${]}} & \boldmath{\textbf{P}~{[}$W${]}} & \textbf{fps}  \\ \midrule
Raspberry Pi 3B+        & IDLE        & 5          & 0.61         & 3.075    & N/A \\
RP3 + Neural Stick 1    & RUNNING     & 5          & 1.2          & 6        & 4    \\
RP3 + Neural Stick 2    & RUNNING     & 5          & 1.12         & 5.6      & 5    \\
Jetson Nano             & IDLE 10W    & 5          & 0.32         & 1.6      & N/A \\
& RUNNING 10W & 5          & 1.96         & 9.8      & 9    \\
& RUNNNING 5W & 5          & 1.4          & 7        & 6    \\
Jetson AGX Xavier       & IDLE 30W    & 19         & 0.35         & 6.65     & N/A \\
& RUNNING 30W & 19         & 0.91         & 17.29    & 30+   \\
& RUNNING 15W & 19         & 0.82         & 15.58    & 28   \\
& RUNNING 10W & 19         & 0.62         & 11.78    & 15   \\
RP3 + Coral Accelerator & MAX         & 5          & 1.40         & 7        & 30+ *  \\
\bottomrule
\end{tabular}
\label{tab:performance}
\end{table*}
\unskip

\subsection{Platform~Implementation}
\begin{figure}[H]
\centering
\includegraphics[width=.482\textwidth]{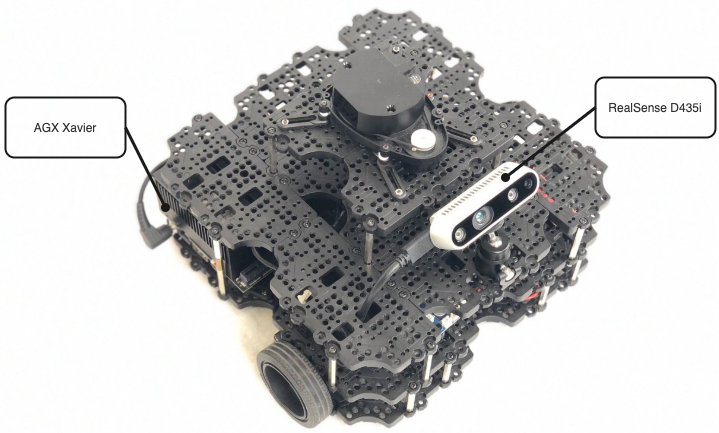}
\caption{TurtleBot3 Waffle platform with NVIDIA Jetson AGX Xavier and Intel RealSense Depth camera D435i~onboard.}
\label{fig:jaffle}
\end{figure}

We tested the proposed cost-effective person-following system in a real environment with the configuration presented in {Table}~\ref{tab:HWcomponents}; the~specifics about its hardware components have been already introduced in Section \ref{subsection:S3_2}. Figure~\ref{fig:jaffle} shows the assembled robot adopted for this~application.

\begin{table}[H]
\caption{{Hardware configuration} adopted for practical simulations. Jetson AGX Xavier and Coral USB Accelerator are used to run the model~onboard.}
\centering
\begin{tabular}{ll}
\toprule
\multicolumn{2}{c}{\textbf{HW Materials}}\\ \midrule
\multicolumn{1}{l}{Robotic Platform}  & TurtleBot3 Waffle Pi \\ \midrule
\multicolumn{1}{l}{Edge AI Device} &  NVIDIA Jetson AGX Xavier\\ \midrule
\multicolumn{1}{l}{RGB-D camera}   & Intel RealSense Depth Camera D435i\\ \bottomrule
\end{tabular}
\label{tab:HWcomponents}
\end{table} 

The tests, performed in a real environment, show robot behavior consistent with expectations. The~person detection algorithm is high-speed and reached a high level of performance.
The network improvement, obtained from the re-training, is considerable (30.09\%), and~this implies optimal real-time results and perfect control of the movements of the robot that follows the~person.

By testing the network on the Jetson Xavier board, we have obtained a high frame rate (30+ fps at the maximum power of the board), as~presented in the Table~\ref{tab:performance}. These results affect the velocity of the detection algorithm that runs in real-time, and~consequently the frequency of the overall control system, running on the Jetson Xavier, which ranges between 18 and 27 Hz.

During the experiments, conducted in the test environment, we have noticed that using all the outputs of the network to modify the control velocity could be counterproductive because it causes a continuous variation of the inputs of the robot control and consequently a not regular movement of the robot. Differently, imposing the updating of the output of the robot command velocity each 2~Hz, the~tests give optimal results: the robot follows the target always in real-time, but~with a notable increase in the smoothness of its movement. It is essential to underline that the implementation of the control adopted has been explicitly designed for the use case taken into account, considering an elderly person as the target and the speed limits of the considered robotic platform: $\pm$0.26 m/s for linear velocity and $\pm$1.8 rad/s for angular velocity. These limits are reasonable for the considered indoor application since walking sessions are usually short and performed with very limited speed and frequent pauses. However, the~control rule can be easily adapted to more performing prototypes if a higher linear speed limit is needed.

During the test phase, the~presented control of the robot has been perfected, in~particular, considering the characteristics of the~platform:
\begin{itemize}
\item[-] The optimal distance limits able to define the three areas of the linear velocity control are defined as $m_{v_{lowlim}}$ = 1.7 m and $m_{v_{uplim}}$ = 1.9 m. In~the range between these two values the robot is in the safe distance zone, so it can only rotate because the linear velocity stays at zero, in~order to avoid the generation of any dangerous situations for the target person.
\item[-] The best linear increment is computed during both the forward and backward movements of the~robot.
\end{itemize}

\begin{figure}[H]
\centering
\subfloat[a][\label{fig:angular_velocity}]
{\includegraphics[width=0.45\textwidth]{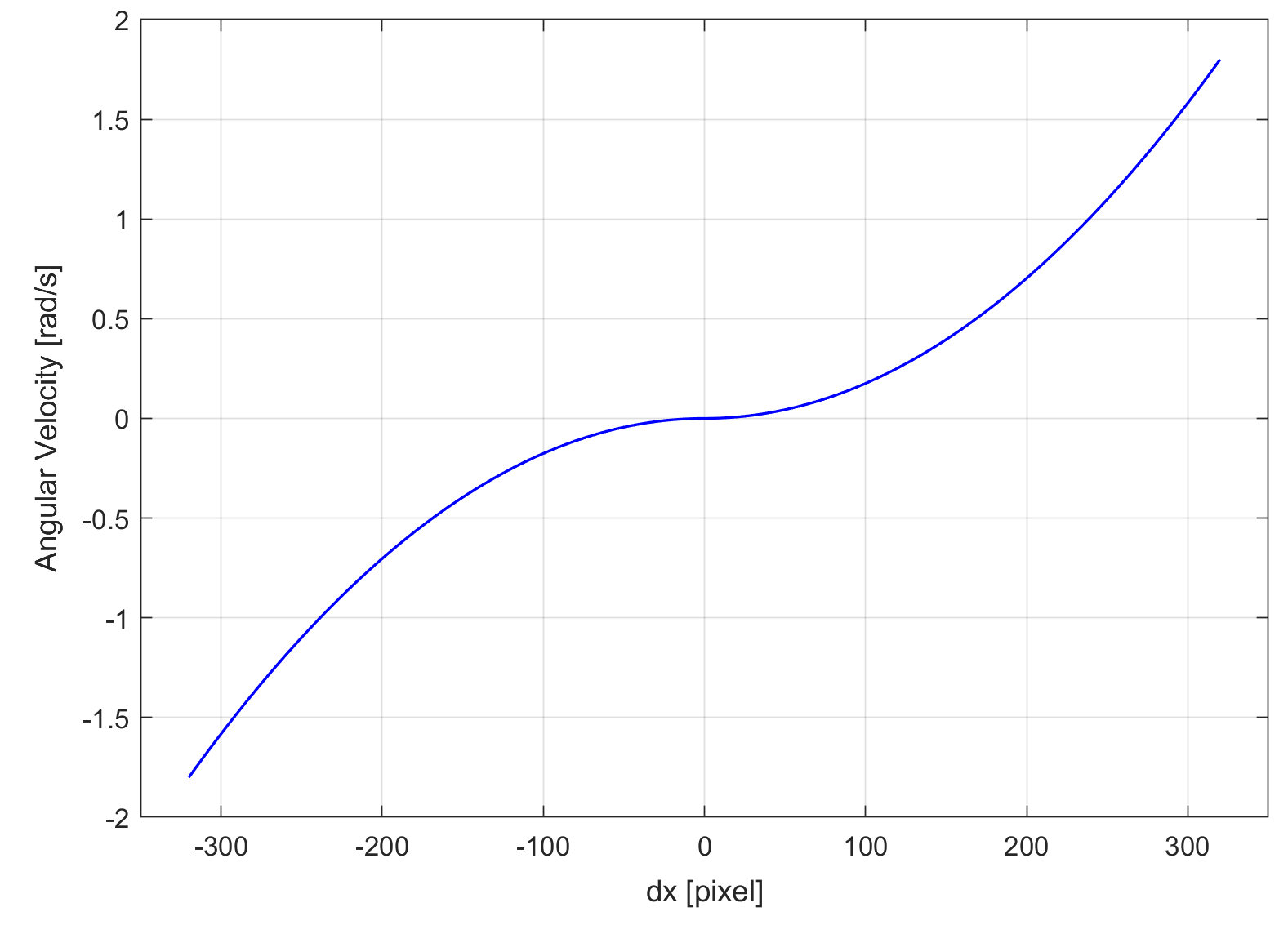}} \quad
\subfloat[b][\label{fig:linear_velocity}]
{\includegraphics[width=0.45\textwidth]{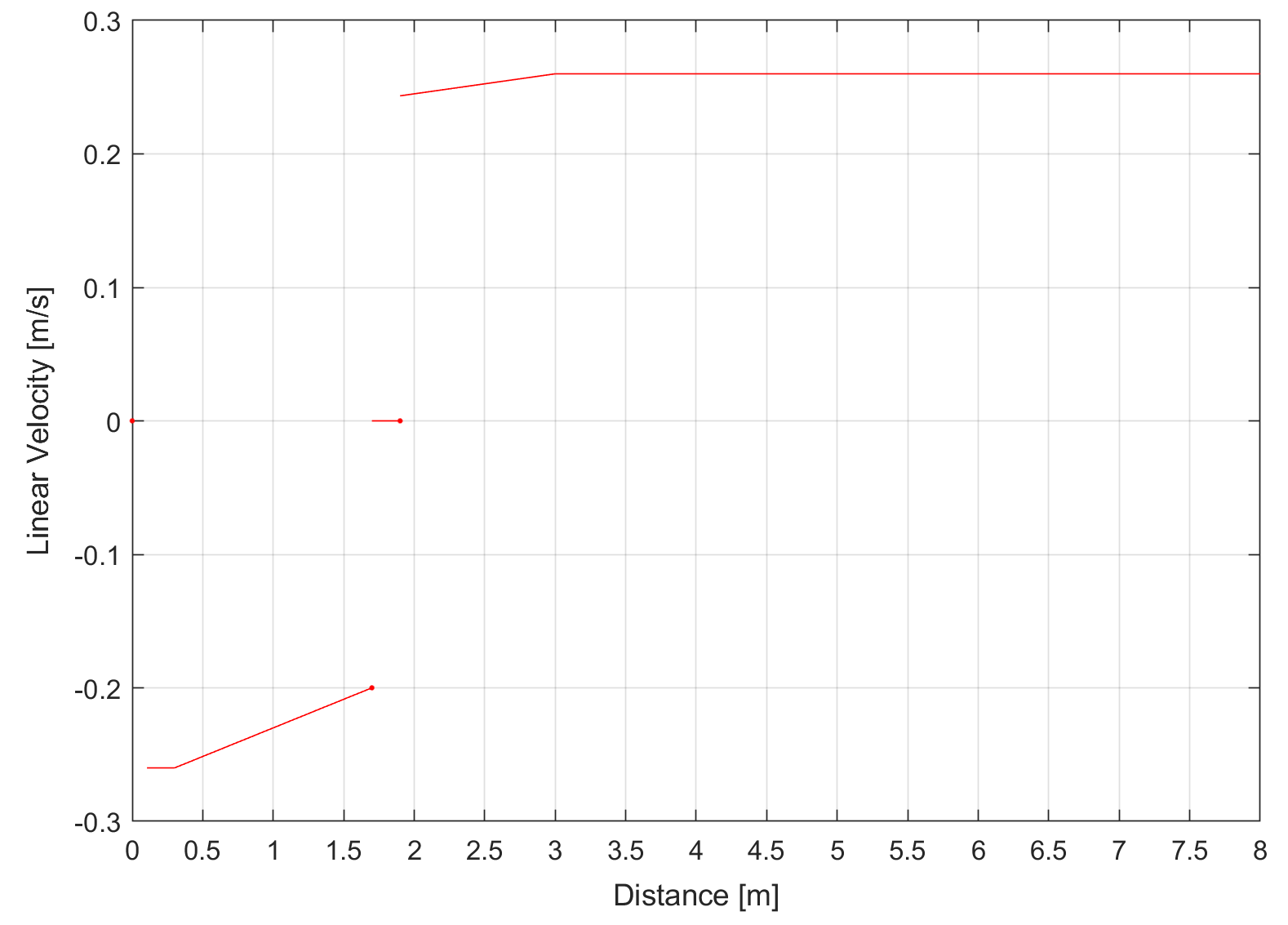}}
\caption{In (\textbf{a}) is the angular velocity control plot, computed considering 640 $\times$ 480 as the resolution of the camera frame and the limit value of the angular velocity $\pm$1.8 rad/s. Instead,~in~(\textbf{b})~is shown the linear velocity control function, computed considering the depth camera range (from 0.105~m to 8~m), the~limit value of the linear velocity of the robotic platform $\pm$0.26~m/s and the safety distance respected inside the interval between 1.7 and 1.9~m.}
\label{fig:velocity_control}
\end{figure}

Thus, the~final obtained values of $m1$, $q1$, $m2$ and $q2$, presented in Section~\ref{sec:methodology}, for~our platform implementation, are reported in Table~\ref{tab:values2}. Moreover, in~Figure~\ref{fig:angular_velocity} the developed angular velocity and linear velocity behaviors are represented while taking into account $max_{vel}$ and $min_{vel}$ of the robotic~platform.

It is important to underline that the initial linear velocity control has been designed with different slopes and without singularities. However, during~the test phase, trouble has been highlighted: the~robot in the restart moving phase proceeded so slowly that it was unable to follow the target without losing it correctly. That is the reason for the introduction of the step singularities, visible on Figure~\ref{fig:linear_velocity}, that, thanks to the small linear velocity of both the robot and the target (the older person), allow one to have a balanced movement and not~jerky.

\begin{table}[H]
\centering
\caption{The values $m1$, $q1$, $m2$, $q2$ identified, after~a test phase, as~the best choice of the linear control algorithm, taking into account the robot adopted in our case study, are here~reported.}
\begin{tabular}{ccc}
\toprule
\textbf{Straight Line} & \textbf{Points} \\ \midrule
\multirow{2}{*}{$1^\circ$: ($m1$, $q1$)
}  &{$P_{1}$} (1 m, 0.23 m/s) \\ \cmidrule{2-2}
&{$P_{2}$} (3 m, 0.26 m/s) \\ \midrule

\multirow{2}{*}{$2^\circ$: ($m2$, $q2$)} &{$P_{1}$} (1 m, $-$0.23 m/s) \\ \cmidrule{2-2}
&{$P_{2}$} (0.3 m, $-$0.26 m/s) \\ \bottomrule
\end{tabular}
\label{tab:values2}
\end{table} 

The overall system has been tested in several environments with different light conditions and target velocities verifying the correctness and completeness of system functionality.
The final result meets the demands of an accurate real-time application; the~robot moves in safety, consistent with the movements of the person, limiting the chances of losing the target to~chase.

We can conclude that we have not realized a simple object tracker, but~a person-following method that focuses on cost-effectiveness, since it cuts unnecessary computations to have a low-cost, functional system. Besides, the~detection is realized at the edge, so the network is optimized to run on neural accelerators. In~this way, it is not necessary to have an expensive computer onboard the robot, which~would imply both the increase of the price and the consumption of additional power.
The reduction in computational cost and power consumption let us  use different types of hardware, presented in Table~\ref{tab:edgeAIcomponents}, associated with their respective performances, reported in Table~\ref{tab:performance}.

\section{Conclusions}
We proposed a cost-effective person-following system for self-sufficient older adults assistance that exploits latest advancements in deep learning optimization techniques and edge AI devices to bring inference directly on the robotic platform with high performance and limited power consumption. We~tested different embedded device configurations, and~we presented a possible practical implementation to realize the suggested system. The~discussed solution is easily replaceable and fully-integrable in pre-existing navigation stacks. Future research may integrate the person-following method with concrete applications and monitoring tools for self-sufficient older~adults.


\section*{acknowledgments}
This work has been developed with the contributions of the Politecnico di Torino Interdepartmental Centre for Service Robotics PIC4SeR (\url{https://pic4ser.polito.it}) and SmartData@Polito (\mbox{\url{https://smartdata.polito.it}}).



\ifCLASSOPTIONcaptionsoff
  \newpage
\fi


\begin{thebibliography}{999}
\providecommand{\natexlab}[1]{#1}

\bibitem[Islam \em{et~al.}(2019)Islam, Hong, and Sattar]{islam2019person}
Islam, M.J.; Hong, J.; Sattar, J.
\newblock Person-following by autonomous robots: A categorical overview.
\newblock {\em  Int. J. Robot.~Res.} {\bf 2019}, {\em
38},~1581--1618. [\href{http://dx.doi.org/10.1177/0278364919881683}{CrossRef}]

\bibitem[wor(2020)]{worldpopulationageing2019}
\newblock {\em World Population Ageing 2019 (ST/ESA/SER.A/444)}; Population Division, Department of Economic and Social Affairs, United Nations: New York, NY, USA, 2020.

\bibitem[LeCun \em{et~al.}(2015)LeCun, Bengio, and Hinton]{lecun2015deep}
LeCun, Y.; Bengio, Y.; Hinton, G.
\newblock Deep learning.
\newblock {\em Nature} {\bf 2015}, {\em 521},~436--444. [\href{http://dx.doi.org/10.1038/nature14539}{CrossRef}] [\href{http://www.ncbi.nlm.nih.gov/pubmed/26017442}{PubMed}]

\bibitem[Redmon and Farhadi(2018)]{redmon2018yolov3}
Redmon, J.; Farhadi, A.
\newblock Yolov3: An incremental improvement.
\newblock {\em arXiv} {\bf 2018}, arXiv:1804.02767.

\bibitem[Pucci \em{et~al.}(2013)Pucci, Marchetti, and
Morin]{pucci2013nonlinear}
Pucci, D.; Marchetti, L.; Morin, P.
\newblock Nonlinear control of unicycle-like robots for person following.
\newblock In~Proceedings~of~the  2013 IEEE/RSJ International Conference on Intelligent Robots and
Systems, Tokyo, Japan,  3--7 November 2013; pp.~3406--3411.

\bibitem[Chung \em{et~al.}(2011)Chung, Kim, Yoo, Moon, and
Park]{chung2011detection}
Chung, W.; Kim, H.; Yoo, Y.; Moon, C.B.; Park, J.
\newblock The detection and following of human legs through inductive
approaches for a mobile robot with a single laser range finder.
\newblock {\em IEEE Trans. Ind. Electron.} {\bf 2011}, {\em
59},~3156--3166. [\href{http://dx.doi.org/10.1109/TIE.2011.2170389}{CrossRef}]

\bibitem[Morales~Saiki \em{et~al.}(2012)Morales~Saiki, Satake, Huq, Glas,
Kanda, and Hagita]{morales2012people}
Morales~Saiki, L.Y.; Satake, S.; Huq, R.; Glas, D.; Kanda, T.; Hagita, N.
\newblock How do people walk side-by-side? Using a computational model of human
behavior for a social robot.
\newblock In~Proceedings~of~the Seventh Annual ACM/IEEE International Conference
on Human-Robot Interaction, Boston, MA, USA, 5--8 March 2012; pp.~301--308.

\bibitem[Cosgun \em{et~al.}(2013)Cosgun, Florencio, and
Christensen]{cosgun2013autonomous}
Cosgun, A.; Florencio, D.A.; Christensen, H.I.
\newblock Autonomous person following for telepresence robots.
\newblock In~Proceedings~of~the  2013 IEEE International Conference on Robotics and Automation, Karlsruhe, Germany, 6--10 May 2013; pp.~4335--4342.

\bibitem[Leigh \em{et~al.}(2015)Leigh, Pineau, Olmedo, and
Zhang]{leigh2015person}
Leigh, A.; Pineau, J.; Olmedo, N.; Zhang, H.
\newblock Person tracking and following with 2d laser scanners.
\newblock  In~Proceedings~of~the 2015 IEEE International Conference on Robotics and Automation
(ICRA), Seattle, WA, USA, 26--30 May 2015; pp.~726--733.

\bibitem[Adiwahono \em{et~al.}(2017)Adiwahono, Saputra, Ng, Gao, Ren, Tan, and
Chang]{adiwahono2017human}
Adiwahono, A.H.; Saputra, V.B.; Ng, K.P.; Gao, W.; Ren, Q.; Tan, B.H.; Chang,
T.
\newblock Human tracking and following in dynamic environment for service
robots.
\newblock In~Proceedings~of~the  TENCON 2017--2017 IEEE Region 10 Conference, Penang, Malaysia, 5--8 November 2017; pp. 3068--3073.

\bibitem[Cen \em{et~al.}(2019)Cen, Huang, Zhong, Peng, and Zou]{cen2019real}
Cen, M.; Huang, Y.; Zhong, X.; Peng, X.; Zou, C.
\newblock Real-time Obstacle Avoidance and Person Following Based on Adaptive
Window Approach.
\newblock In~Proceedings~of~the  2019 IEEE International Conference on Mechatronics and Automation
(ICMA), Tianjin, China, 4--7 August 2019; pp.~64--69.

\bibitem[Jung \em{et~al.}(2012)Jung, Yi, et~al.]{jung2012control}
Jung, E.J.; Yi, B.J.; Yuta, S.
\newblock Control algorithms for a mobile robot tracking a human in front.
\newblock In~Proceedings~of~the  2012 IEEE/RSJ International Conference on Intelligent Robots and
Systems, Vilamoura, Portugal, 7--12 October 2012; pp.~2411--2416.

\bibitem[Cai and Matsumaru(2014)]{cai2014human}
Cai, J.; Matsumaru, T.
\newblock Human detecting and following mobile robot using a laser range
sensor.
\newblock {\em J. Robot. Mechatron.} {\bf 2014}, {\em
26},~718--734. [\href{http://dx.doi.org/10.20965/jrm.2014.p0718}{CrossRef}]

\bibitem[Koide and Miura(2016)]{koide2016identification}
Koide, K.; Miura, J.
\newblock Identification of a specific person using color, height, and gait
features for a person following robot.
\newblock {\em Robot Auton. Syst.} {\bf 2016}, {\em 84},~76--87. [\href{http://dx.doi.org/10.1016/j.robot.2016.07.004}{CrossRef}]

\bibitem[Brookshire(2010)]{brookshire2010person}
Brookshire, J.
\newblock Person following using histograms of oriented gradients.
\newblock {\em Int. J. Soc. Robot.} {\bf 2010}, {\em
2},~137--146. [\href{http://dx.doi.org/10.1007/s12369-010-0046-y}{CrossRef}]

\bibitem[Satake \em{et~al.}(2012)Satake, Chiba, and Miura]{satake2012sift}
Satake, J.; Chiba, M.; Miura, J.
\newblock A SIFT-based person identification using a distance-dependent
appearance model for a person following robot.
\newblock  In~Proceedings~of~the 2012 IEEE International Conference on Robotics and Biomimetics
(ROBIO), Guangzhou, China, 11--14 December 2012; pp.~962--967.

\bibitem[Satake \em{et~al.}(2013)Satake, Chiba, and Miura]{satake2013visual}
Satake, J.; Chiba, M.; Miura, J.
\newblock Visual person identification using a distance-dependent appearance
model for a person following robot.
\newblock {\em Int. J. Autom. Comput.} {\bf 2013},
{\em 10},~438--446. [\href{http://dx.doi.org/10.1007/s11633-013-0740-y}{CrossRef}]

\bibitem[Chen \em{et~al.}(2017{\natexlab{a}})Chen, Sahdev, and
Tsotsos]{chen2017integrating}
Chen, B.X.; Sahdev, R.; Tsotsos, J.K.
\newblock Integrating stereo vision with a CNN tracker for a person-following
robot.
\newblock In \emph{International Conference on Computer Vision Systems}; Springer: Berlin/Heidelberg, Germany, 2017; pp.~300--313.  


\bibitem[Chen \em{et~al.}(2017{\natexlab{b}})Chen, Sahdev, and
Tsotsos]{chen2017person}
Chen, B.X.; Sahdev, R.; Tsotsos, J.K.
\newblock Person following robot using selected online ada-boosting with stereo
camera.
\newblock  In~Proceedings~of~the 2017 14th Conference on Computer and Robot Vision (CRV), Edmonton, AB, Canada, 16--19 May 2017; pp.~48--55.

\bibitem[Wang \em{et~al.}(2018)Wang, Zhang, Wang, and Hu]{wang2018person}
Wang, X.; Zhang, L.; Wang, D.; Hu, X.
\newblock Person detection, tracking and following using stereo camera.
\newblock  In~Proceedings~of~the Ninth International Conference on Graphic and Image Processing
(ICGIP 2017), Qingdao, China, 14--17 October 2017; p. 106150D.

\bibitem[Doisy \em{et~al.}(2012)Doisy, Jevtic, Lucet, and
Edan]{doisy2012adaptive}
Doisy, G.; Jevtic, A.; Lucet, E.; Edan, Y.
\newblock Adaptive person-following algorithm based on depth images and
mapping.
\newblock  In~Proceedings~of~the IROS Workshop on Robot Motion Planning, Vilamoura, Portugal, 7--12~October 2012.

\bibitem[Basso \em{et~al.}(2013)Basso, Munaro, Michieletto, Pagello, and
Menegatti]{basso2013fast}
Basso, F.; Munaro, M.; Michieletto, S.; Pagello, E.; Menegatti, E.
\newblock Fast and robust multi-people tracking from RGB-D data for a mobile
robot. In {\em Intelligent Autonomous Systems 12}; Springer: Berlin/Heidelberg, Germany, 2013; pp.
265--276.

\bibitem[Munaro \em{et~al.}(2013)Munaro, Basso, Michieletto, Pagello, and
Menegatti]{munaro2013software}
Munaro, M.; Basso, F.; Michieletto, S.; Pagello, E.; Menegatti, E.
\newblock A software architecture for RGB-D people tracking based on ROS
framework for a mobile robot. In {\em Frontiers of Intelligent Autonomous
Systems}; Springer: Berlin/Heidelberg, Germany, 2013; pp. 53--68.

\bibitem[Do and Lin(2015)]{do2015embedded}
Do, M.Q.; Lin, C.H.
\newblock Embedded human-following mobile-robot with an RGB-D camera.
\newblock  In~Proceedings~of~the 2015 14th IAPR International Conference on Machine Vision
Applications (MVA), Tokyo, Japan, 18--22 May 2015; pp.~555--558.

\bibitem[Ren \em{et~al.}(2016)Ren, Zhao, Qi, and Li]{ren2016real}
Ren, Q.; Zhao, Q.; Qi, H.; Li, L.
\newblock Real-time target tracking system for person-following robot.
\newblock  In~Proceedings~of~the 2016 35th Chinese Control Conference (CCC), Chengdu, China, 27--29 July 2016; pp. 6160--6165.

\bibitem[Mi \em{et~al.}(2016)Mi, Wang, Ren, and Hou]{mi2016system}
Mi, W.; Wang, X.; Ren, P.; Hou, C.
\newblock A system for an anticipative front human following robot.
\newblock  In~Proceedings~of~the International Conference on Artificial
Intelligence and Robotics and the International Conference on Automation,
Control and Robotics Engineering, Kitakyushu, Japan, 12--15 July 2016; pp.~1--6.

\bibitem[Gupta \em{et~al.}(2016)Gupta, Kumar, Behera, and
Subramanian]{gupta2016novel}
Gupta, M.; Kumar, S.; Behera, L.; Subramanian, V.K.
\newblock A novel vision-based tracking algorithm for a human-following mobile
robot.
\newblock {\em IEEE Trans. Syst. Man Cybern. Syst.}
{\bf 2016}, {\em 47},~1415--1427. [\href{http://dx.doi.org/10.1109/TSMC.2016.2616343}{CrossRef}]

\bibitem[Masuzawa \em{et~al.}(2017)Masuzawa, Miura, and
Oishi]{masuzawa2017development}
Masuzawa, H.; Miura, J.; Oishi, S.
\newblock Development of a mobile robot for harvest support in greenhouse
horticulture---Person following and mapping.
\newblock In~Proceedings~of~the  2017 IEEE/SICE International Symposium on System Integration (SII), Taipei, Taiwan, 11--14 December 2017; pp.~541--546.

\bibitem[Chi \em{et~al.}(2017)Chi, Wang, and Meng]{chi2017gait}
Chi, W.; Wang, J.; Meng, M.Q.H.
\newblock A gait recognition method for human following in service robots.
\newblock {\em IEEE~Trans. Syst. Man Cybern. Syst.}
{\bf 2017}, {\em 48},~1429--1440. [\href{http://dx.doi.org/10.1109/TSMC.2017.2660547}{CrossRef}]

\bibitem[Jiang \em{et~al.}(2018)Jiang, Yao, Hong, Li, Su, and
Kuc]{jiang2018classification}
Jiang, S.; Yao, W.; Hong, Z.; Li, L.; Su, C.; Kuc, T.Y.
\newblock A classification-lock tracking strategy allowing a person-following
robot to operate in a complicated indoor environment.
\newblock {\em Sensors} {\bf 2018}, {\em 18},~3903. [\href{http://dx.doi.org/10.3390/s18113903}{CrossRef}] [\href{http://www.ncbi.nlm.nih.gov/pubmed/30424577}{PubMed}]

\bibitem[Chen(2018)]{chen2018folo}
Chen, E.
\newblock ``FOLO'': A Vision-Based Human-Following Robot.
\newblock  In~{Proceedings~of~the 2018 3rd International Conference on Automation, Mechanical Control
and Computational Engineering (AMCCE 2018)}, Dalian, China, 12--13 May 2018; Atlantis Press: Beijing, China, 2018.


\bibitem[Yang and Song(2019)]{yang2019control}
Yang, C.A.; Song, K.T.
\newblock Control Design for Robotic Human-Following and Obstacle Avoidance
Using an RGB-D Camera.
\newblock  In~Proceedings~of~the 2019 19th International Conference on Control, Automation and
Systems (ICCAS), Jeju, Korea, 15--18 October 2019; pp.~934--939.

\bibitem[Alvarez-Santos \em{et~al.}(2012)Alvarez-Santos, Pardo, Iglesias,
Canedo-Rodriguez, and Regueiro]{alvarez2012feature}
Alvarez-Santos, V.; Pardo, X.M.; Iglesias, R.; Canedo-Rodriguez, A.; Regueiro,
C.V.
\newblock Feature analysis for human recognition and discrimination:
Application to a person-following behaviour in a mobile robot.
\newblock {\em Robot. Auton. Syst.} {\bf 2012}, {\em
60},~1021--1036. [\href{http://dx.doi.org/10.1016/j.robot.2012.05.014}{CrossRef}]

\bibitem[Susperregi \em{et~al.}(2013)Susperregi, Mart{\'\i}nez-Otzeta,
Ansuategui, Ibarguren, and Sierra]{susperregi2013rgb}
Susperregi, L.; Mart{\'\i}nez-Otzeta, J.M.; Ansuategui, A.; Ibarguren, A.;
Sierra, B.
\newblock RGB-D, laser and thermal sensor fusion for people following in a
mobile robot.
\newblock {\em Int. J. Adv. Robot. Syst.} {\bf 2013},
{\em 10},~271. [\href{http://dx.doi.org/10.5772/56123}{CrossRef}]

\bibitem[Wang \em{et~al.}(2017)Wang, Su, Shi, Liu, and Miro]{wang2017real}
Wang, M.; Su, D.; Shi, L.; Liu, Y.; Miro, J.V.
\newblock Real-time 3D human tracking for mobile robots with multisensors.
\newblock  In~Proceedings~of~the 2017 IEEE International Conference on Robotics and Automation
(ICRA), Singapore, 29 May--3 June 2017; pp.~5081--5087.

\bibitem[Hu \em{et~al.}(2013)Hu, Wang, and Ho]{hu2013design}
Hu, J.S.; Wang, J.J.; Ho, D.M.
\newblock Design of sensing system and anticipative behavior for human
following of mobile robots.
\newblock {\em IEEE Trans. Ind. Electron.} {\bf 2013}, {\em
61},~1916--1927. [\href{http://dx.doi.org/10.1109/TIE.2013.2262758}{CrossRef}]

\bibitem[Dalal and Triggs(2005)]{dalal2005histograms}
Dalal, N.; Triggs, B.
\newblock Histograms of oriented gradients for human detection.
\newblock  In~Proceedings~of~the 2005 IEEE Computer Society Conference on Computer Vision and Pattern
Recognition (CVPR'05), San Diego, CA, USA, 20--25 June 2005; pp.~886--893.

\bibitem[Redmon \em{et~al.}(2016)Redmon, Divvala, Girshick, and
Farhadi]{redmon2016you}
Redmon, J.; Divvala, S.; Girshick, R.; Farhadi, A.
\newblock You only look once: Unified, real-time object detection.
\newblock  In~Proceedings~of~the IEEE Conference on Computer Vision and Pattern
Recognition, Las Vegas, NV, USA, 26 June--1 July 2016; pp.~779--788.

\bibitem[Viola and Jones(2001)]{viola2001rapid}
Viola, P.; Jones, M.
\newblock Rapid object detection using a boosted cascade of simple features.
\newblock  In~Proceedings~of~the IEEE Computer Society Conference on Computer
Vision and Pattern Recognition  (CVPR 2001), Kauai, HI, USA, 8--14 December 2001; pp.511--518.

\bibitem[Felzenszwalb \em{et~al.}(2008)Felzenszwalb, McAllester, and
Ramanan]{felzenszwalb2008discriminatively}
Felzenszwalb, P.; McAllester, D.; Ramanan, D.
\newblock A discriminatively trained, multiscale, deformable part model.
\newblock  In~Proceedings~of~the 2008 IEEE Conference on Computer Vision and Pattern Recognition, Anchorage, AK, USA, 23--28 June 2008; pp.~1--8.

\bibitem[Girshick \em{et~al.}(2014)Girshick, Donahue, Darrell, and
Malik]{girshick2014rich}
Girshick, R.; Donahue, J.; Darrell, T.; Malik, J.
\newblock Rich feature hierarchies for accurate object detection and semantic
segmentation.
\newblock  In~Proceedings~of~the IEEE Conference on Computer Vision and Pattern
Recognition, Columbus, OH, USA, 23--28 June 2014; pp.~580--587.

\bibitem[Girshick(2015)]{girshick2015fast}
Girshick, R.
\newblock Fast r-cnn.
\newblock  In~Proceedings~of~the IEEE International Conference on Computer Vision, Santiago, Chile, 7--13 December 2015; pp.~1440--1448.

\bibitem[Ren \em{et~al.}(2015)Ren, He, Girshick, and Sun]{ren2015faster}
Ren, S.; He, K.; Girshick, R.; Sun, J.
\newblock Faster r-cnn: Towards real-time object detection with region proposal
networks.
\newblock In~Proceedings~of~ twenty-ninth Conference on Neural Information Processing Systems, Montreal, QC, Canada, 7--12 December 2015; pp. 91--99.

\bibitem[Liu \em{et~al.}(2016)Liu, Anguelov, Erhan, Szegedy, Reed, Fu, and
Berg]{Liu-2016}
Liu, W.; Anguelov, D.; Erhan, D.; Szegedy, C.; Reed, S.; Fu, C.Y.; Berg, A.C.
\newblock SSD: Single Shot MultiBox Detector.
\newblock {\em Lect. Notes Comput. Sci.} {\bf 2016}, 21--37. [\href{http://dx.doi.org/10.1007/978-3-319-46448-0_2}{CrossRef}]

\bibitem[Redmon and Farhadi(2017)]{redmon2017yolo9000}
Redmon, J.; Farhadi, A.
\newblock YOLO9000: Better, faster, stronger.
\newblock  In~Proceedings~of~the IEEE Conference on Computer Vision and Pattern
Recognition, Honolulu, HI, USA, 21--26 July 2017; pp.~7263--7271.

\bibitem[Bochkovskiy \em{et~al.}(2020)Bochkovskiy, Wang, and
Liao]{bochkovskiy2020yolov4}
Bochkovskiy, A.; Wang, C.Y.; Liao, H.Y.M.
\newblock YOLOv4: Optimal Speed and Accuracy of Object Detection.
\newblock {\em arXiv}~{\bf 2020}, arXiv:2004.10934.

\bibitem[Mittal(2019)]{mittal2019survey}
Mittal, S.
\newblock A Survey on optimized implementation of deep learning models on the
NVIDIA Jetson platform.
\newblock {\em J. Syst. Archit.} {\bf 2019}, {\em 97},~428--442. [\href{http://dx.doi.org/10.1016/j.sysarc.2019.01.011}{CrossRef}]

\bibitem[Xu \em{et~al.}(2017)Xu, Amaro, Caulfield, Falcao, and
Moloney]{xu2017classify}
Xu, X.; Amaro, J.; Caulfield, S.; Falcao, G.; Moloney, D.
\newblock Classify 3D voxel based point-cloud using convolutional neural
network on a neural compute stick.
\newblock  In~Proceedings~of~the 2017 13th International Conference on Natural Computation, Fuzzy
Systems and Knowledge Discovery (ICNC-FSKD), Guilin, China, 29--31 July 2017; pp.~37--43.

\bibitem[Kang \em{et~al.}(2018)Kang, Kang, Kang, Yoo, and Ha]{kang2018joint}
Kang, D.; Kang, D.; Kang, J.; Yoo, S.; Ha, S.
\newblock Joint optimization of speed, accuracy, and energy for embedded image
recognition systems.
\newblock In~Proceedings~of~the  2018 Design, Automation \& Test in Europe Conference \& Exhibition
(DATE), Dresden, Germany, 19--23 March 2018; pp.~715--720.

\bibitem[Cao \em{et~al.}(2018)Cao, Liu, Lasang, and Shen]{cao2018detecting}
Cao, S.; Liu, Y.; Lasang, P.; Shen, S.
\newblock Detecting the objects on the road using modular lightweight network.
\newblock {\em arXiv} {\bf 2018}, arXiv:1811.06641.

\bibitem[Yang \em{et~al.}(2018)Yang, Ren, Zhang, Xie, Ren, Li, and
Zhang]{yang2018hybrid}
Yang, T.; Ren, Q.; Zhang, F.; Xie, B.; Ren, H.; Li, J.; Zhang, Y.
\newblock Hybrid camera array-based uav auto-landing on moving ugv in
gps-denied environment.
\newblock {\em Remote Sens.} {\bf 2018}, {\em 10},~1829. [\href{http://dx.doi.org/10.3390/rs10111829}{CrossRef}]

\bibitem[Mazzia \em{et~al.}(2020)Mazzia, Khaliq, Salvetti, and
Chiaberge]{mazzia2020real}
Mazzia, V.; Khaliq, A.; Salvetti, F.; Chiaberge, M.
\newblock Real-Time Apple Detection System Using Embedded Systems With Hardware
Accelerators: An Edge AI Application.
\newblock {\em IEEE Access} {\bf 2020}, {\em 8},~9102--9114. [\href{http://dx.doi.org/10.1109/ACCESS.2020.2964608}{CrossRef}]

\bibitem[Long \em{et~al.}(2016)Long, Zhu, Wang, and Jordan]{long2016deep}
Long, M.; Zhu, H.; Wang, J.; Jordan, M.I.
\newblock Deep Transfer Learning with Joint Adaptation Networks. {\em arXiv}~{\bf 2016}, arXiv:1605.06636.

\bibitem[Kuznetsova \em{et~al.}(2018)Kuznetsova, Rom, Alldrin, Uijlings,
Krasin, Pont-Tuset, Kamali, Popov, Malloci, Duerig,
et~al.]{kuznetsova2018open}
Kuznetsova, A.; Rom, H.; Alldrin, N.; Uijlings, J.; Krasin, I.; Pont-Tuset, J.;
Kamali, S.; Popov, S.; Malloci,~M.; Duerig, T.; et al.
\newblock The open images dataset v4: Unified image classification, object
detection, and visual relationship detection at scale.
\newblock {\em arXiv} {\bf 2018}, arXiv:1811.00982.

\bibitem[Wojke \em{et~al.}(2017)Wojke, Bewley, and Paulus]{wojke2017simple}
Wojke, N.; Bewley, A.; Paulus, D.
\newblock Simple online and realtime tracking with a deep association metric.
\newblock  In~Proceedings~of~the 2017 IEEE International Conference on Image Processing (ICIP), Beijing, China, 17--20~September 2017; pp.~3645--3649.

\bibitem[Chen \em{et~al.}(2018)Chen, Ai, Zhuang, and Shang]{Chen2018RealTimeMP}
Chen, L.; Ai, H.; Zhuang, Z.; Shang, C.
\newblock Real-Time Multiple People Tracking with Deeply Learned Candidate
Selection and Person Re-Identification. In~Proceedings~of~the 2018 IEEE International Conference on Multimedia and Expo
(ICME), San Diego, CA, USA, 23--27 July 2018; pp.~1--6.

\bibitem[Mitzel and Leibe(2011)]{mitzel2011real}
Mitzel, D.; Leibe, B.
\newblock Real-time multi-person tracking with detector assisted structure
propagation.
\newblock  In~Proceedings~of~the 2011 IEEE International Conference on Computer Vision Workshops
(ICCV Workshops), Barcelona, Spain, 6--13 November 2011; pp.~974--981.

\bibitem[Tan \em{et~al.}(2018)Tan, Sun, Kong, Zhang, Yang, and
Liu]{tan2018survey}
Tan, C.; Sun, F.; Kong, T.; Zhang, W.; Yang, C.; Liu, C.
\newblock A survey on deep transfer learning.
\newblock In  \emph{International Conference on Artificial Neural Networks}; Springer: Berlin/Heidelberg, Germany, 2018; pp.~270--279.

\bibitem[Polyak(1964)]{polyak1964some}
Polyak, B.T.
\newblock Some methods of speeding up the convergence of iteration methods.
\newblock {\em USSR Comput. Math. Math. Phys.} {\bf
1964}, {\em 4},~1--17. [\href{http://dx.doi.org/10.1016/0041-5553(64)90137-5}{CrossRef}]

\end{thebibliography}

\vfill\break

\begin{IEEEbiography}[{\includegraphics[width=1in,height=1.30in,clip,keepaspectratio]{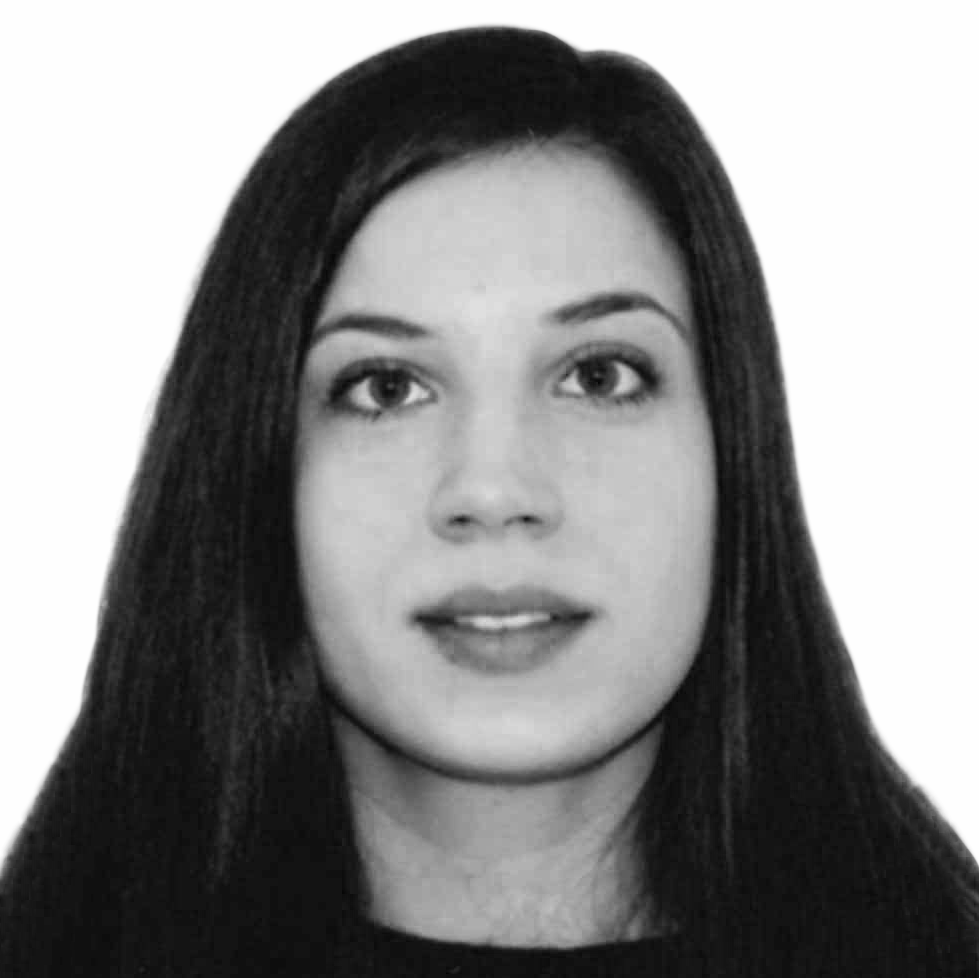}}]{Anna Boschi} is a researcher at PIC4SeR - PoliTO Interdepartmental Centre for Service Robotics (\url{https://pic4ser.polito.it/}).
She received a master's degree in Mechatronics Engineering from Politecnico di Torino with the thesis “Person tracking methodologies and algorithms in service robotic applications” carried out at PIC4SeR. Currently, she is focusing on service robotics for indoor applications combining computer vision and Machine Learning algorithms with autonomous navigation methodologies.
\end{IEEEbiography}

\begin{IEEEbiography}[{\includegraphics[width=1in,height=1.30in,clip,keepaspectratio]{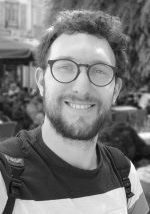}}]{Francesco Salvetti} is currently a Ph.D. student in Electrical, Electronics and Communications Engineering in collaboration with the two interdepartmental centers PIC4SeR (\url{https://pic4ser.polito.it/}) and Smart Data (\url{https://smartdata.polito.it/}) at Politecnico di Torino, Italy. He received his Bachelor's Degree in Electronic Engineering§ in 2017 and his Master’s Degree in Mechatronics Engineering in 2019 at Politecnico di Torino. He is currently working on Machine Learning applied to Computer Vision and Image Processing in robotics applications.
\end{IEEEbiography}

\begin{IEEEbiography}[{\includegraphics[width=1in,height=1.25in,clip,keepaspectratio]{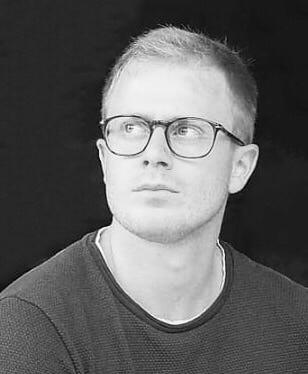}}]{Vittorio Mazzia} is a PhD student in Electrical, Electronics and Communications Engineering working with the two Interdepartmental Centres PIC4SeR (\url{https://pic4ser.polito.it/}) and SmartData (\url{https://smartdata.polito.it/}). He received a master's degree in Mechatronics Engineering from the Politecnico di Torino, presenting a thesis entitled "Use of deep learning for automatic low-cost detection of cracks in tunnels," developed in collaboration with the California State University. His current research interests involve deep learning applied to different tasks of computer vision, autonomous navigation for service robotics, and reinforcement learning. Moreover, making use of neural compute devices (like Jetson Xavier, Jetson Nano, Movidius Neural Stick) for hardware acceleration, he is currently working on machine learning algorithms and their embedded implementation for AI at the edge.
\end{IEEEbiography}

\begin{IEEEbiography}[{\includegraphics[width=1in,height=1.25in,clip,keepaspectratio]{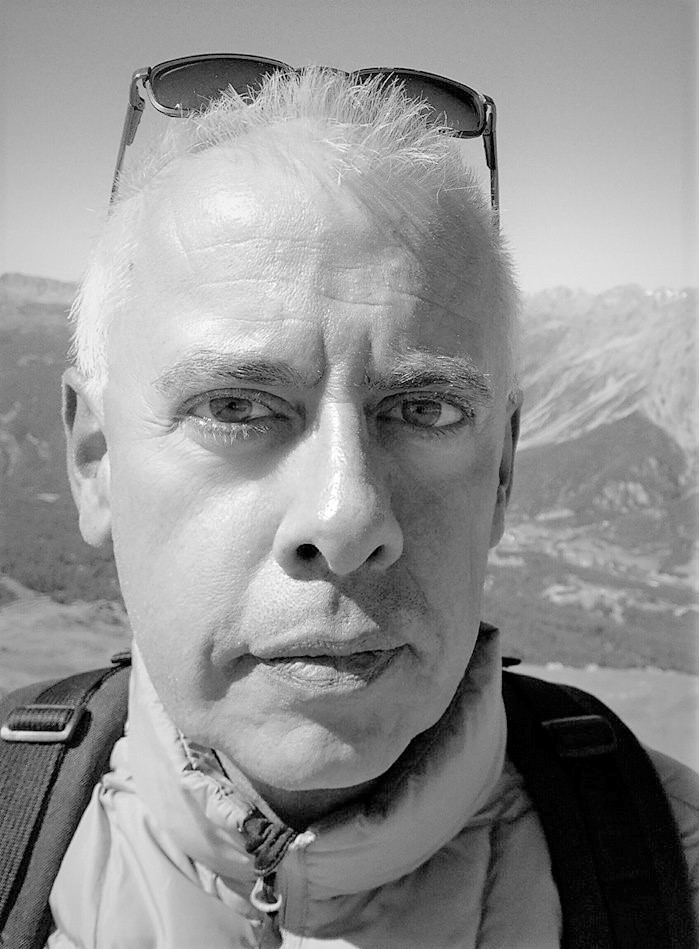}}]{Marcello Chiaberge} is currently Associate Professor within the Department of Electronics and Telecommunications, Politecnico di Torino, Turin, Italy. He is also the Co-Director of the Mechatronics Lab, Politecnico di Torino
(\url{www.lim.polito.it}), Turin, and the Director and the Principal Investigator of the new Centre for Service Robotics (PIC4SeR, \url{https://pic4ser.polito.it/}), Turin. He has authored more than 100 articles accepted in international conferences and journals,
and he is the coauthor of nine international patents. His research interests include
hardware implementation of neural networks and fuzzy systems and the design and implementation of reconfigurable real-time computing architectures. \end{IEEEbiography}
\vfill

\end{document}